\documentclass[letterpaper]{article} 
\usepackage[]{aaai2026}  
\usepackage{times}  
\usepackage{helvet}  
\usepackage{courier}  
\usepackage[hyphens]{url}  
\usepackage{graphicx} 
\usepackage{natbib}  
\usepackage{caption} 
\usepackage{algorithm}
\usepackage{algorithmic}

\usepackage{multirow}
\usepackage[table,xcdraw]{xcolor}
\usepackage{booktabs}
\usepackage{amsmath}

\usepackage{subcaption}
\usepackage{newfloat}
\usepackage{listings}
\DeclareCaptionStyle{ruled}{labelfont=normalfont,labelsep=colon,strut=off} 
\floatstyle{ruled}
\newfloat{listing}{tb}{lst}{}
\floatname{listing}{Listing}

\usepackage{colortbl} 
\usepackage[most]{tcolorbox}
\newenvironment{myexampleblock}[1]{%
    \tcolorbox[enhanced,%
    noparskip,
    title=#1,
    colback=blue!10,
    colframe=blue!75!black]}%
    {\endtcolorbox}

\title{Can LLMs Truly Embody Human Personality? Analyzing AI and Human Behavior Alignment  in Dispute Resolution}

\author{Deuksin Kwon\textsuperscript{\rm 1,2}\equalcontrib, 
Kaleen Shrestha\textsuperscript{\rm 1}\equalcontrib, 
Bin Han\textsuperscript{\rm 1,2},
Spencer Lin\textsuperscript{\rm 1},
James Hale\textsuperscript{\rm 1,2}, \\
Jonathan Gratch\textsuperscript{\rm 1,2},
Maja Matarić\textsuperscript{\rm 1},
Gale M. Lucas\textsuperscript{\rm 1,2}
}

\affiliations{
    \textsuperscript{\rm 1}Department of Computer Science, University of Southern California\\
    \textsuperscript{\rm 2}USC Institute for Creative Technologies

    \{deuksink, kshresth, binhan, linspenc, jahale, mataric\}@usc.edu\\
    \{gratch, lucas\}@ict.usc.edu}
\usepackage{bibentry}

\begin{document}

\maketitle

\begin{abstract}
Large language models (LLMs) are increasingly used to simulate human behavior in social settings such as legal mediation, negotiation, and dispute resolution. However, it remains unclear whether these simulations reproduce the personality–behavior patterns observed in humans. Human personality, for instance, shapes how individuals navigate social interactions, including strategic choices and behaviors in emotionally charged interactions. This raises the question: \textit{Can LLMs, when prompted with personality traits, reproduce personality-driven differences in human conflict behavior?} To explore this, we introduce an evaluation framework that enables direct comparison of human-human and LLM-LLM behaviors in dispute resolution dialogues with respect to Big Five Inventory (BFI) personality traits. This framework provides a set of interpretable metrics related to strategic behavior and conflict outcomes. We additionally contribute a novel dataset creation methodology for LLM dispute resolution dialogues with matched scenarios and personality traits with respect to human conversations. Finally, we demonstrate the use of our evaluation framework with three contemporary closed-source LLMs and show significant divergences in how personality manifests in conflict across different LLMs compared to human data, challenging the assumption that personality-prompted agents can serve as reliable behavioral proxies in socially impactful applications. Our work highlights the need for psychological grounding and validation in AI simulations before real-world use. 
\end{abstract}


\section{Introduction}
The ability to simulate human behavior in high-stakes interpersonal contexts—such as negotiation and dispute resolution—has growing societal relevance. As large language models (LLMs) are increasingly deployed in socially impactful settings, from conflict resolution coaching to AI-assisted decision-making \cite{lin2024imbue, ashtiani2023news, li2025text, kwon-etal-2025-astra}, it is essential to assess whether they reflect core psychological factors that guide human behavior. One such factor is personality, which plays a central role in shaping how people navigate interpersonal conflict \cite{antonioni1998relationship, bell1977personality, wood2008predicting}.

Conflict resolution often involves dynamic and strategic decisions—such as whether to assert, accommodate, or withdraw—made under conditions of uncertainty and evolving interpersonal dynamics~\cite{ross1991barriers}. Among the various factors influencing these behaviors, personality traits such as agreeableness, neuroticism, and openness to experience systematically predict individual differences in conflict-related behavior~\cite{antonioni1998relationship, tehrani2020personality}. Accounting for personality provides important insights into the social-cognitive mechanisms that drive interpersonal dynamics.

Accordingly, a multitude of studies have examined personality’s role in conflict resolution; however, most have relied on static questionnaires or simplified decision tasks~\cite{sternberg1984styles, wood2008predicting}, limiting insight into how personality is behaviorally expressed in complex, unfolding, goal-directed interactions~\cite{baumeister2007psychology}. Furthermore, research has rarely moved beyond human-only paradigms or LLM-only simulations to examine whether personality-linked behaviors remain consistent across systems. Given the growing use of LLMs to simulate social behavior, this raises a critical empirical question: \textit{Can LLMs reproduce personality-driven differences in human conflict behavior when guided by personality prompts?}

While recent studies have begun exploring personality-prompted LLMs for generating individualized dialogue~\cite{serapiogarcía2023personality, jiang2023evaluating}, the assumption that these models serve as psychologically accurate proxies remains largely untested. The behavioral fidelity of LLMs, particularly in emotionally charged settings, has not been rigorously validated. This gap limits our ability to trust LLMs in applications where misalignment with human behavior could have social/ethical consequences.

To address this gap, we introduce an evaluation framework for systematically assessing how personality manifests in human-human and LLM-LLM conflict resolution dialogues. Leveraging the KObe DISpute corpus (KODIS) dataset~\cite{hale2025kodis}, which features multi-issue, turn-based negotiations embedded in disputes, we construct a parallel set of LLM dialogues by prompting LLMs with matched scenarios and Big Five Inventory (BFI) personality profiles. This framework facilitates fine-grained comparisons of strategic behaviors and conflict outcomes between human and LLM interactions.

Methodologically, we introduce an experimental paradigm that mirrors a human–human dialogue corpus by matching agents on personality and fixing scenarios, enabling one-to-one human–LLM comparisons, and we propose a novel, generalizable framework to evaluate LLMs’ behavioral alignment with human psychological constructs. Empirically, applying this framework to recent LLMs reveals consistent mismatches: in humans, neuroticism is the strongest predictor of strategic outcomes, whereas models show stronger effects for extraversion and agreeableness and exhibit strategic behaviors that load on a broader set of personality factors. Among LLMs, Claude and Gemini align more closely with human strategic metrics than GPT-4o mini, indicating partial convergence with human patterns. Our framework, simulation code, and supplementary materials are publicly available \footnote{\url{https://github.com/DSincerity/Personality-LLM-BehavAlign-Dispute}}.

These findings challenge the growing assumption that personality-prompted LLMs can reliably serve as human proxies in social applications. As LLMs become more integrated in socially consequential domains, our work underscores the need for psychological grounding, interpretability, and behavioral validation. Ultimately, our framework provides a path toward more human-aligned, socially responsible AI systems.

\section{Background and Related Work}
To inform our research, we investigate prior work on personality and conflict, as well as personality prompting of LLMs. 

\subsection{Personality and Conflict Resolution Strategy}
Personality traits are influential factors in how individuals perceive and respond to interpersonal conflict~\cite{wood2008predicting,tehrani2020personality,antonioni1998relationship}.
Early studies identified links between the BFI personality traits and conflict resolution styles. For example, ~\citet{jones1982personality} and ~\citet{bell1977personality} found that preferences for competing, accommodating, and avoiding conflict resolution styles vary based on personality.
~\citet{wood2008predicting} further showed that extraversion and agreeableness significantly predict conflict style based on the Thomas–Kilmann model. Participants with higher extraversion or agreeableness are more likely to adopt collaborative or accommodating approaches.
~\citet{kaushal2006role} examined how personality-related traits, specifically emotional intelligence, self-monitoring, and cultural values, influence conflict resolution strategy. The study found that collectivism and interpersonal harmony were associated with preferences for harmony-preserving strategies. ~\citet{park2007personality} used questionnaires and found that extraversion and agreeableness were linked to cooperative conflict strategies, especially when the other party was cooperative. Neurotic individuals, in contrast, favored avoidant or dominating approaches regardless of the partner’s behavior. 
~\citet{graziano1996perceiving} examined how agreeableness shapes conflict resolution preferences and behavior. Agreeable individuals consistently favored negotiation and disengagement over power assertion. 

\subsection{LLMs with Personality}
Recent advances in LLMs have enabled the simulation of complex human social behaviors~\cite{zhou2023sotopia, parksocialsimulacra}. These advances have enabled research into simulations between multiple LLMs, in domains such as negotiation and other decision-making contexts~\cite{kwon-etal-2025-astra, abdelnabi2024cooperation, xie2024can}.

To enrich social behavior, recent studies have explored prompting LLMs to emulate personality profiles, particularly those based on BFI ~\cite{serapiogarcía2023personality, jiang2023evaluating, sorokovikova2024llms, gui2023challenge, han2025can}. Incorporating such traits allows LLMs to better reflect personality differences, enriching the psychological realism and diversity of multi-LLM interactions.

\citet{serapiogarcía2023personality} and \citet{jiang2023evaluating} proposed prompting frameworks for eliciting and modeling personality traits in LLMs. Other works show LLMs can exhibit or be edited toward trait-consistent responses \cite{sorokovikova2024llms, mao2024editing}. Building on these capabilities, recent work has employed personality-conditioned LLM simulations. Notably, \citet{huang-hadfi-2024-personality} used BFI-prompted LLMs in negotiation settings and observed trait-driven differences in strategy and preferences. This line of research highlights LLMs’ potential for simulating psychologically grounded behavior at scale.

However, much of the past work assumes that LLMs prompted with specific traits behave analogously to humans with similar dispositions. This assumption, however, remains largely untested, especially in complex social contexts such as conflict and negotiation, where behavior reflects internal traits and interpersonal dynamics. Prior studies have focused on trait expression, not on whether LLM behaviors align with human behavior. Our study systematically compares personality–behavior relationships in LLMs and humans under matched negotiation scenarios, highlighting key differences in behavior.
\begin{figure*}[th]
    \centering
    \includegraphics[width=0.80\linewidth]{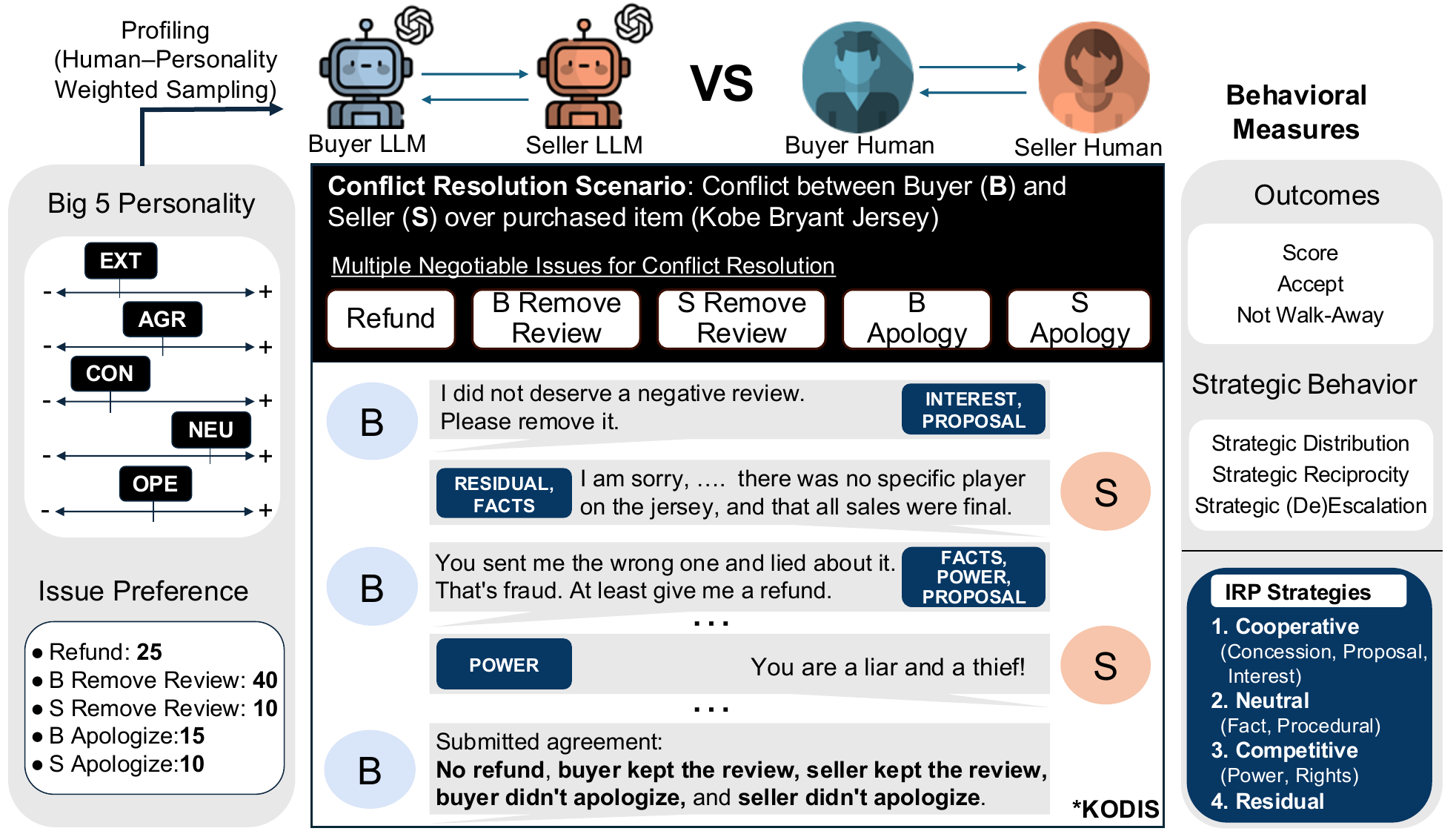}
    \caption{Overview of the conflict resolution scenario, an example dialogue from the KODIS dataset, the profiling setup for LLM simulation, and the behavioral evaluation measures.}
    \label{fig:main_methods}
\vspace{-2ex}
\end{figure*}
\section{Methodology} \label{sec:methodology}
To explore personality in conflict resolution dialogue and compare human and LLM behavior, we introduce the datasets and conflict resolution behavioral measures to investigate our proposed research questions.
\subsection{KODIS: Human vs. Human Dataset}
The \emph{KObe DISpute corpus} (KODIS) is a role-play dispute resolution dataset collected from crowd-sourced participants on Prolific by \citet{hale2025kodis}. It consists of extended dialogues between two participants (out of 4,061) with Buyer or Seller roles. As shown in Figure~\ref{fig:main_methods}, the scenario is an emotionally charged dispute over a jersey purchased online for a sick nephew. The two parties have different perspectives and strong negative emotions towards each other. They then negotiate a resolution based on their preference over a set number of issues: (1) a full refund, (2) removal of negative reviews, or (3) a formal apology. We excluded human–AI conversations from the original dataset and used a subset of 248 human–human dialogues with complete personality information from both participants. Each analysis used a filtered subset with missing values removed as appropriate.

\subsection{Behavioral Measures}
We define two types of behavioral measures: (1) final outcomes, and (2) strategic behaviors.

\subsubsection{Final Outcomes}\label{sec:action_outcome}
We focus on three final outcome variables of dispute conversation, corresponding to well-established negotiation measures \cite{kelley1996classroom}.

\begin{enumerate}
    \item[(1)] \textit{Score}: The participant’s final payoff, computed as the inner product of the agreed allocation and the participant’s issue preferences.
    \item[(2)] \textit{Accept}: Whether the participant accepted an offer.
   \item[(3)] \textit{Not Walk-Away}: Whether the participant chose to stay in the negotiation rather than walking away. Staying reflects a basic strategic commitment to reach an agreement.
\end{enumerate}

\subsubsection{Strategic Behaviors}
To investigate strategic behaviors, we present four metrics based on the Interests-Rights-Power (IRP) framework~\cite{ury1988getting}. This framework categorizes utterances according to whether they appeal to interest, assert rights, or leverage power, offering a useful lens for understanding how individuals navigate conflict. Figure~\ref{fig:main_methods} illustrates IRP annotations within a conflict dialogue, showing strategy shifts. The IRP framework defines eight utterance types, as shown in Table~\ref{tab:irp-strategies}.

\begin{table*}[t]
\centering
\begin{tabular} {p{1.7cm} p{14.9cm}}
\toprule
\textbf{Strategy} & \textbf{Examples and Definitions} \\
\midrule
\emph{Cooperative} & 
\textbf{Proposal}: Concrete ideas for resolving the conflict; 
\textbf{Concession}: Willingness to change an initial position; 
\textbf{Interests}: Statements of needs or concerns; 
\textbf{Positive Expectations}: Optimistic outlook or recognition of common goals. \\

\emph{Neutral} & 
\textbf{Facts}: Information sharing or clarification; 
\textbf{Procedural}: Statements about conversation structure or norms. \\

\emph{Competitive} & 
\textbf{Power}: Threats or coercive moves; 
\textbf{Rights}: Appeals to rules, norms, or fairness. \\

\emph{Residual} & 
\textbf{Residual}: If an utterance does not fit into any of the above categories. \\
\bottomrule
\end{tabular}
\caption{IRP strategy types and example speech acts.}
\label{tab:irp-strategies}
\vspace{-2ex}
\end{table*}
In prior work, BFI personality traits have been linked to conflict styles \cite{tehrani2020personality}. We examine whether participants’ use of IRP strategies—particularly the most relevant in conflict, \emph{Cooperative} and \emph{Competitive} strategies \cite{brett1998breaking}—is associated with personality. Cooperative mirroring aids resolution \cite{kelley1996classroom}, while competitive mirroring can push conflict into destructive spirals \cite{brett1998breaking}. Thus, we analyze whether participants reciprocate their partner’s \emph{Cooperative} or \emph{Competitive} IRP strategies in the immediate turn, and how these patterns relate to personality. Additionally, we examine escalation and de-escalation, defined as competitive responses to non-competitive moves (and vice versa). Reciprocity and escalation are well-studied concepts in negotiation and conflict research \cite{putnam1982reciprocity, zartman2005dynamics}. Speakers may use these strategically—to assert dominance or to reduce tension and refocus the negotiation \cite{brett1998breaking}. Accordingly, we define the following four metrics:

\begin{itemize}
    \item[(1)] \textbf{IRP Ratio}: The relative frequency of competitive (\emph{C}) or cooperative (\emph{Co}) strategies used by the speaker (\emph{S}):
    \[
    \text{IRP}_{\text{ratio}}^X = \frac{N_S^X}{N_S^\text{all strategies}}, \quad X \in \{C, Co\}
    \]

    \item[(2)] \textbf{IRP Reciprocity}: The proportion of speaker turns that match the partner’s preceding strategy ($X_P$):
    \[
    \text{IRP}_{\text{recip}}^X = \frac{N_S^{X=X_P}}{N_P^X}, \quad X \in \{C, Co\}
    \]

    \item[(3)] \textbf{Escalation Ratio}: The rate at which the speaker responds competitively to non-competitive partner turns ($NC_P$):
    \[
    \text{Escalation} = \frac{N_S^{NC_P \rightarrow C}}{N_P^{NC}}
    \]

    \item[(4)] \textbf{De-escalation Ratio}: The rate at which the speaker responds non-competitively to competitive partner turns:
    \[
    \text{De\text{-}escalation} = \frac{N_S^{C_P \rightarrow NC}}{N_P^C}
    \]
\end{itemize}

\noindent
Here, $N_S^X$ denotes the number of speaker turns using strategy $X$, and $N_S^{Y_P \rightarrow X}$ refers to speaker turns using $X$ in response to the partner’s preceding strategy $Y$. We define \emph{C} and \emph{Co} as competitive and cooperative strategies, respectively, and \emph{NC} as non-competitive strategies (i.e., \emph{Cooperative}, \emph{Neutral}, and \emph{Residual}).

\subsection{L2L: LLM-to-LLM Simulation for Parallel Dataset Construction}
We construct diverse personality profiles and generate corresponding prompts to create LLMs with distinct personality traits. We build the LLM-to-LLM (L2L) dataset, consisting of LLM-based simulations, and analyze their behaviors in conflict resolution. We use OpenAI GPT-4o mini \cite{gpt4o}, Anthropic Claude Sonnet 3.7 \cite{claudesonnet37}, and Gemini 2.0 Flash \cite{gemini2flash}, hereafter GPT-4, Claude, and Gemini. We obtain IRP annotations for the L2L dataset using the same procedure as KODIS. We use the default hyperparameters for all LLMs (temperature of 1) to test the zero-shot learning capabilities of these models.

\subsubsection{LLM Personality Profile} 
Each LLM is assigned a BFI personality profile ($\{P_{AGR}$, $P_{EXT}$, $P_{CON}$, $P_{NEU}$, $P_{OPE}\}$), following the validated personality-prompt design of ~\citet{huang-hadfi-2024-personality}. This approach produces results highly consistent with IPIP tests, ensuring reliable personality manipulations. Prompt variation is minimized to assess if LLMs can inherently represent human personality traits.  

To enable fair comparisons with human data, personality profiles are sampled from the empirical distribution of human BFI traits. Each trait uses a six-point polarity–degree scale (e.g., $\{P_{\text{AGR}}^{++}, P_{\text{EXT}}^{+++}, …\}$).

Building on this, each LLM is assigned a personalized issue importance profile and negotiates more assertively on issues it deems more important. For the \textit{Apology}, importance is weighted by LLM agreeableness based on regression results from human data ($B$=2.13, \textit{p}=.02). Other issues are assigned importance values at random. This personality-informed prioritization introduces psychologically meaningful variation for fair comparison with human behavior.

To generate personality prompts, we use 70 pairs of bipolar adjectives empirically associated with the BFI \cite{goldberg1992development, serapiogarcía2023personality}. Each LLM receives 15 adjectives (three per trait), modified to reflect trait intensity: “very” for high, “a bit” for low, and no modifier for medium. We prompt LLMs with their assigned profile prompt throughout the simulation.

\subsubsection{LLM Simulation} 

\begin{table*}[thb!]
\centering
\small
\renewcommand{\arraystretch}{0.95}
\resizebox{0.75\linewidth}{!}{%
\begin{tabular}{@{}cllll@{}}
\toprule
    & \multicolumn{4}{c}{Significant Independent Variables (IVs) - Beta Coefficient}  \\ \cmidrule(l){2-5} 
\multirow{-2}{*}{\begin{tabular}[c]{@{}c@{}}Dependent \\ Variables (DVs)\end{tabular}}  & \multicolumn{1}{c|}{GPT-4}     & \multicolumn{1}{c|}{Claude }      & \multicolumn{1}{c|}{Gemini}  & \multicolumn{1}{c}{KODIS}     \\ \midrule
\rowcolor[HTML]{ECF4FF} 
\multicolumn{1}{c|}{\cellcolor[HTML]{ECF4FF}Score}   & \multicolumn{1}{l|}{\cellcolor[HTML]{ECF4FF}\begin{tabular}[c]{@{}l@{}}S-EXT (B=1.67**)\\ S-AGR (B=-4.38***)\\ POS (B=16.85***)\end{tabular}} & \multicolumn{1}{l|}{\cellcolor[HTML]{ECF4FF}\begin{tabular}[c]{@{}l@{}}S-AGR (B=-2.50***)\\ P-EXT (B=-1.42*)\\ P-AGR (B=3.05***)\\ POS (B=-12.05***)\end{tabular}} & \multicolumn{1}{l|}{\cellcolor[HTML]{ECF4FF}\begin{tabular}[c]{@{}l@{}}S-AGR (B=-4.48***)\\ S-CON (B=1.72*)\\ S-OPE (B=1.94*)\\ POS (B=-5.31***)\end{tabular}} & POS (B=-3.21*)  \\
\multicolumn{1}{c|}{Accept}     & \multicolumn{1}{l|}{POS (B=-0.22*)}  & \multicolumn{1}{l|}{\begin{tabular}[c]{@{}l@{}}S-EXT (B=-0.17**)\\ P-EXT (B=0.17**)\\ POS (B=-0.47***)\end{tabular}}     & \multicolumn{1}{l|}{}     & \begin{tabular}[c]{@{}l@{}}S-NEU (B=-0.26*)\\ P-NEU (B=0.27*)\\ POS (B=0.49***)\end{tabular} \\
\rowcolor[HTML]{ECF4FF} 
\multicolumn{1}{c|}{\cellcolor[HTML]{ECF4FF}\begin{tabular}[c]{@{}c@{}}Not \\ Walk-Away\end{tabular}} & \multicolumn{1}{l|}{\cellcolor[HTML]{ECF4FF}\begin{tabular}[c]{@{}l@{}}S-OPE (B=-0.18*)\\ P-OPE (B=0.18*)\\ POS (B=0.94***)\end{tabular}}    & \multicolumn{1}{l|}{\cellcolor[HTML]{ECF4FF}}      & \multicolumn{1}{l|}{\cellcolor[HTML]{ECF4FF}\begin{tabular}[c]{@{}l@{}}S-NEU (B=-0.18*)\\ P-NEU (B=0.18*)\end{tabular}}     &   \\ \hline
\multicolumn{5}{c}{Coefficients (B) are reported. *, **, *** indicate p $<$ .05, .01, and .001, respectively.}  \\ \hline\bottomrule
\end{tabular}}
\caption{Summary of significant regression results for personality predictors across LLM (L2L) and human (KODIS) datasets, for conflict resolution action and outcomes. S and P refer to Self and Partner, respectively.}
\label{tab:regression_results}
\vspace{-2ex}
\end{table*}

Using the approach described above, we configure two LLMs—one acting as the Buyer and the other as the Seller—and simulate the conflict scenario from KODIS. As shown in Table~\ref{fig:main_methods}, they negotiate over three core concerns (refund, negative reviews, apology) and must ultimately reach agreement across five issues.

At each turn, LLMs respond to the opponent’s previous utterance by making an offer (\texttt{SUBMIT}), responding (\texttt{ACCEPT}/\texttt{REJECT}), or continuing the dialogue. When negotiation reaches deadlock, LLMs can choose to \texttt{WALK-AWAY}.

The dialogue ends when one of the LLMs chooses to \texttt{ACCEPT} an offer or walk away. The negotiation is considered unsuccessful if neither LLM reaches an agreement within a predefined length limit (\texttt{NO AGREEMENT}).

We first run 500 simulations with GPT-4 and then include additional models (250 simulations for each) to demonstrate the framework’s flexibility. While simulation counts varied for practical reasons, each provided statistically valid estimates. The L2L Prompt and example can be found in the supplementary material.

\subsection{Comparison with Regression Analysis}
To examine how personality traits influence negotiation behavior, we treat the above proposed metrics as dependent variables (DVs). Based on each DV’s data type—continuous or binary (0 or 1)—we applied linear regression or logistic regression, respectively. As independent variables, we included the LLM's own BFI personality traits and those of the partner. Position (e.g., Buyer or Seller) was included as a control variable, using effect coding (Buyer=–1 and Seller=1) (See the supplementary material for model details). These metrics support systematic comparison between humans and LLMs.
\section{Results and Discussion}\label{sec:results}
This section presents results addressing how personality traits shape outcomes and strategic behavior in human conflict dialogues, and whether personality-prompted LLMs align with human behavior. We analyze findings from the KODIS and L2L datasets. Overall, our results reveal divergences between human and LLM behaviors, challenging assumptions of LLM alignment. For all regression findings, we report significant results with $p<.05$.

\subsection{Effects of Personality Traits on Actions and Outcomes}
\label{sec:results-conflict-behavior}

We examined how personality traits influence three key negotiation outcomes—\textit{score, offer acceptance}, and \textit{not walk-away} behavior—comparing humans (KODIS) with LLMs to evaluate the degree of behavioral alignment (See Table~\ref{tab:regression_results}). Full results can be found in the supplementary material.

\paragraph{\textit{Score}}
Human scores showed no significant associations with any personality trait, suggesting that outcomes were shaped more by interactional dynamics than by stable dispositions. Conversely, all LLMs exhibited clear trait-based effects. Higher agreeableness predicted lower scores. For GPT-4, agreeableness negatively influenced outcomes across both roles (Buyer: $B$=$–2.62$, $p$=$.001$; Seller: $B$=$–3.32, p$=$.000$). Model-specific patterns also emerged: GPT-4 performed better when more extraverted ($B$=$1.67$, $p$=$.005$); Gemini benefited from higher conscientiousness and openness; and Claude achieved higher scores when paired with introverted but agreeable partners, reflecting responsiveness to partner traits.

\paragraph{\textit{Accept}}
Humans showed role-contingent effects of neuroticism. Individuals high in neuroticism were significantly less likely to accept offers ($B$=$–0.26$, $p$=$.026$), while negotiating with neurotic partners increased acceptance likelihood ($B$=$0.27$, $p$=$.025$). This asymmetry suggests reluctance to commit under emotional sensitivity and possible appeasement of emotionally volatile counterparts. Among LLMs, GPT-4 and Gemini showed no significant trait effects, but Claude was more likely to accept when introverted and paired with an extroverted partner—a pattern not mirrored in humans.

\paragraph{\textit{Not Walk-Away}}
For humans, no personality traits significantly predicted the likelihood of staying in the negotiation, suggesting that such decisions were driven more by contextual factors. In contrast, GPT-4 and Gemini were significantly less likely to remain engaged when personality mismatches with their partners were high, indicating a unique LLM sensitivity to interpersonal dissimilarity that may lead to premature disengagement under trait misalignment.

\begin{table*}[t]
\centering
\small
\renewcommand{\arraystretch}{0.95}
\resizebox{\linewidth}{!}{%
\begin{tabular}{cllllll}
\hline
        & \multicolumn{6}{c}{Significant Independent Variables (IVs) - Beta Coefficient}\\ \cline{2-7} 
\multirow{-2}{*}{Dataset} & \multicolumn{1}{c}{Cooperative Ratio}      & \multicolumn{1}{c}{Competitive Ratio}       & \multicolumn{1}{c}{Cooperative Reciprocity}    & \multicolumn{1}{c}{Competitive Reciprocity}     & \multicolumn{1}{c}{Escalation Ratio}  & \multicolumn{1}{c}{De-escalation Ratio}       \\ \hline
\rowcolor[HTML]{DAE8FC} 
{\color[HTML]{000000} GPT-4} & {\color[HTML]{000000} \begin{tabular}[c]{@{}l@{}}P-EXT (B=0.49*)\\ POS (B=1.56***)\end{tabular}} & {\color[HTML]{000000} \begin{tabular}[c]{@{}l@{}}S-EXT (B=0.95*)\\ S-NEU (B=0.88*)\\ P-EXT (B=1.01*) \\ P-NEU (B=0.89*)\end{tabular}} & {\color[HTML]{000000} \begin{tabular}[c]{@{}l@{}}S-AGR (B=1.10***)\\ P-EXT (B=-0.89*)\\ POS (B=-6.58***)\end{tabular}} & {\color[HTML]{000000} \begin{tabular}[c]{@{}l@{}}S-EXT (B=1.40*)\\ S-AGR (B=-1.46*)\\ S-NEU (B=1.41*) \\ P-CON (B=-1.41*) \\ POS (B=13.27***)\end{tabular}} & {\color[HTML]{000000} \begin{tabular}[c]{@{}l@{}}S-AGR (B=-1.37**)\\ P-EXT (B=1.56**)\\ POS (B=11.21***)\end{tabular}} & {\color[HTML]{000000} \begin{tabular}[c]{@{}l@{}}S-AGR (B=1.83***)\\ POS (B=-7.57***)\end{tabular}} \\
Gemini        & POS (B=1.88***) & \begin{tabular}[c]{@{}l@{}}S-EXT (B=1.07*)\\ P-EXT (B=1.09*)\end{tabular}       & \begin{tabular}[c]{@{}l@{}}S-EXT (B=3.08***)\\ S-AGR (B=1.51**)\\ POS (B=-17.18***)\end{tabular}     & POS (B=7.75***)      & POS (B=3.30**)      & \begin{tabular}[c]{@{}l@{}}S-EXT (B=4.13**)\\ P-EXT (B=2.85*)\\ POS (B=-19.33***)\end{tabular}      \\
\rowcolor[HTML]{DAE8FC} 
Claude & POS (B=3.24***) && POS (B=-7.02***)    & POS (B=13.20***)     & \begin{tabular}[c]{@{}l@{}}S-AGR (B=-2.45***)\\ S-NEU (B=1.02*)\\ POS (B=6.85**)\end{tabular}        &  \\
KODIS   &        & \begin{tabular}[c]{@{}l@{}}S-EXT (B=1.74*)\\ P-EXT (B=1.80*)\\ P-CON (B=-1.77*)\end{tabular}      & \begin{tabular}[c]{@{}l@{}}S-OPE (B=4.02*)\\ P-CON (B=-4.98**)\\ POS (B=-3.79*)\end{tabular}& POS (B=8.36**)       & POS (B=8.25***)     &  \\ \hline
\multicolumn{7}{c}{Coefficients (B) are reported. *, **, *** indicate p $<$ .05, .01, and .001, respectively.}  \\ \hline
\end{tabular}}
\caption{Regression results for the IRP-related dependent variables: IRP Ratio, IRP Reciprocity Ratio, and Escalation/De-escalation Ratio. S, P, and POS refer to Self, Partner, and Position, respectively. Full results for individual IRP strategies for ratio and reciprocity metrics can be found in the supplementary material.}
\label{tab:irp_regression_results} 
\vspace{-2ex}
\end{table*}

\subsubsection{Role-Dependent Personality Effects and Human–LLM Alignment}
We also examined whether personality effects vary by negotiation role (Buyer vs. Seller) and found partial alignment between LLMs and humans in role-contingent behavior. While both GPT-4 and humans showed role-dependent effects of agreeableness in offer acceptance, only LLMs showed personality effects on final scores. Full results can be found in the supplementary material.

\subsection{Effects of Personality Traits on Strategic Behavior}
\label{sec:results-strategic-behavior}
We examined how personality relates to the frequency and reciprocity of \emph{Competitive} and \emph{Cooperative} IRP strategies, as well as the frequency of escalation and de-escalation responses in the KODIS and L2L datasets (Table \ref{tab:irp_regression_results}). Full results can be found in the supplementary material.



\subsubsection{Personality Effects on IRP Ratio}
As presented in Table \ref{tab:irp_regression_results} for L2L, several significant personality effects emerged for the frequency of \emph{Cooperative} and \emph{Competitive} strategies.

For \emph{Cooperative} strategies, GPT-4 showed a significant effect of partner extraversion ($B$=0.49, $p$=.031), which did not appear in KODIS. Gemini and Claude showed no significant predictors, similar to KODIS. For \emph{Competitive} strategies, GPT-4 aligned more closely with KODIS, showing significant effects for self- and partner-extraversion ($B$=0.95, $p$=.025; $B$=1.01, $p$=.017), as well as neuroticism ($B$=0.88, $p$=.030; $B$=0.89, $p$=.028). Gemini also overlapped with KODIS, with significant effects for both self- and partner-extraversion ($B$=1.07, $p$=.043; $B$=1.09, $p$=.041). Claude showed no overlap. These results suggest that Gemini better captures personality effects for \emph{Cooperative} and \emph{Competitive} strategy frequencies. 

Overall, personality effects on IRP strategy for GPT-4 and Claude show low alignment with KODIS, aside from a partial match observed for Cooperative strategies, whereas Gemini showed a higher overlap for both strategy types.

\begin{figure*}[t!]
    \centering
    \includegraphics[width=1\linewidth]{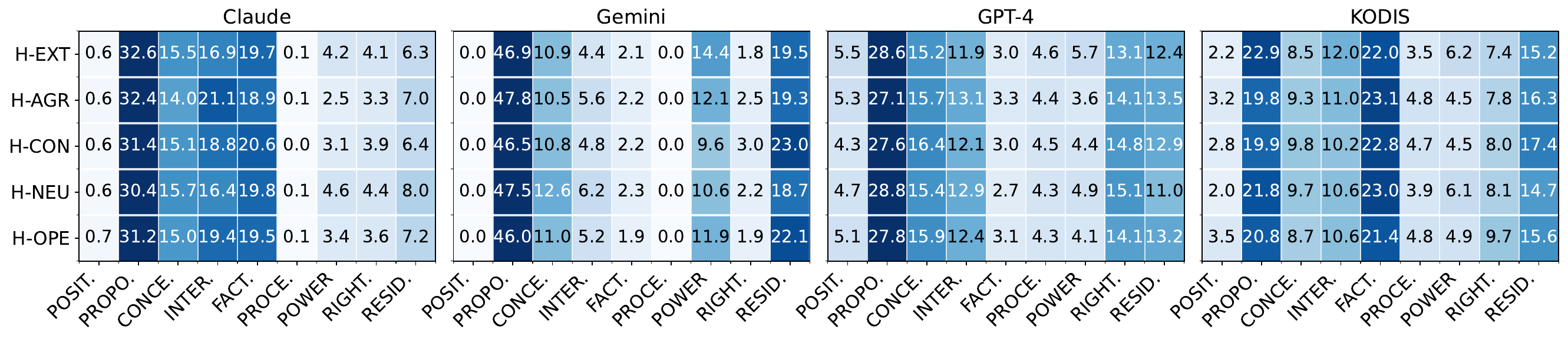}
    \caption{IRP strategy heatmap by personality traits across LLMs and human dialogues; rows sum to 100\% (IRP strategy distribution per trait). H denotes “high,” and the x-axis labels represent the first five letters of each strategy listed in Table \ref{tab:irp-strategies}.}
    \label{fig:irp_personality_heatmap}
    \vspace{-2ex}
\end{figure*}
Figure~\ref{fig:irp_personality_heatmap} shows IRP strategy distributions across five personality traits. Trait-driven variation across traits is minimal for both humans and LLMs, suggesting limited sensitivity to personality differences.

Humans rely most on \emph{Facts} and show the most balanced distribution, reflecting flexible, context-sensitive behavior. LLMs, in contrast, favor \emph{Proposal} and use \emph{Concession} more, indicating a more transactional style.

Model-specific patterns also vary. Claude most closely resembles humans, with higher use of \emph{Facts}. Gemini shows the most skewed distribution, with elevated \emph{Residual} and \emph{Power}, and no use of \emph{Positive Expectations} or \emph{Procedural} moves. GPT-4 is more balanced but consistently uses more \emph{Power}.

These findings suggest that while LLMs tend to over-rely on transactional strategies, they differ in how rigidly or assertively those strategies are applied, highlighting the need for greater adaptability and nuance.

\subsubsection{Personality Effects on IRP Reciprocity}
For reciprocity of \emph{Cooperative} strategies in the LLMs (Table~\ref {tab:irp_regression_results}), we find no overlap with the KODIS dataset for GPT-4, with significant effects of partner-extraversion ($B$=$-0.89, p$=$.013$) and self-agreeableness ($B$=$1.10, p$=$.000$). The same goes for Gemini, where there is also a significant effect of self-agreeableness ($B$=$1.51, p$=$.002$), and additionally for self-extraversion ($B$=$3.08, p$=$.000$). Claude shows no significant personality effects. For \emph{Competitive} strategies, no overlap with KODIS is observed for GPT-4; in fact, multiple personality traits—self-extraversion, self-agreeableness, self-neuroticism, and partner-conscientiousness—significantly influence reciprocity ($B$=$1.40, p$=$.049$; $B$=$-1.46, p$=$.034$; $B$=$1.41, p$=$.042$; $B$=$-1.41, p$=$.038$), whereas there were no significant personality effects in KODIS. Gemini and Claude both align with KODIS, indicating no significant personality effects. These findings demonstrate that Gemini and Claude align with human results for competitive reciprocity, while all three LLMs diverge for cooperative reciprocity, suggesting that competitive reciprocity patterns are more strongly associated with personality.

\subsubsection{Personality Effects on (De)Escalation}
For escalation response frequency, only Gemini aligns with KODIS, with no significant personality effects. GPT-4 and Claude both have significant personality effects, with self-agreeableness in common ($B$=$-1.37, p$=$.004$ and $B$=$-2.45, p$=$.000$). GPT-4 additionally has significant effects for partner-extraversion ($B$=$1.56, p$=$.002$) and Claude additionally has significant effects for self-neuroticism ($B$=$1.02, p$=$.024$).

For de-escalation, Claude aligns with KODIS with no significant personality effects. Gemini has significant personality effects for self- and partner-extraversion ($B$=$4.13, p$=$.002$) and ($B$=$2.85, p$=$.023$). GPT-4 has significant effects for self-agreeableness ($B$=$-1.83, p$=$.000$). These results show that Claude was aligned with KODIS on de-escalation, and Gemini was aligned with KODIS on escalation response frequencies, with GPT-4 being least aligned for both measures.

\subsubsection{Comparison of IRP Reciprocity and (De)Escalation Frequencies in L2L vs. KODIS}
\begin{figure}[t!]
    \centering
    \includegraphics[width=1\linewidth]{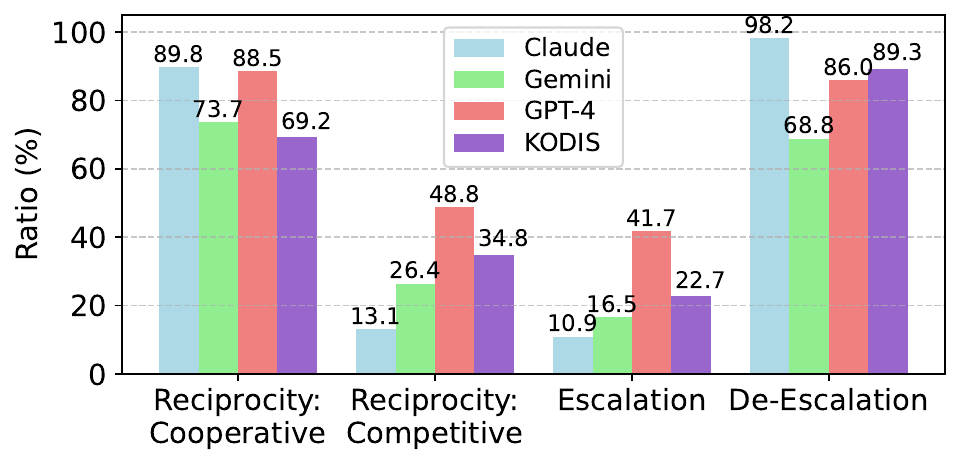}
    \caption{Comparison of frequencies of Cooperative/Competitive reciprocity and (De) Escalation for KODIS and L2L.}
    \label{fig:reciprocity_kodis_a2a}
    \vspace{-2ex}
\end{figure}

Figure~\ref{fig:reciprocity_kodis_a2a} compares the frequency of IRP Reciprocity (Cooperative and Competitive) and (De) Escalation responses, for humans (KODIS) and LLMs, aggregated across traits due to minimal variation.

LLMs more consistently reciprocate \emph{Cooperative} strategies, whereas humans show more flexible reciprocity, including greater \emph{Competitive} reciprocity. Among LLMs, GPT-4 displays particularly strong responses to competitive moves, suggesting heightened sensitivity to adversarial cues.

Escalation patterns further distinguish models from humans. GPT-4 tends to escalate more readily, while Claude strongly favors de-escalation, with minimal escalation throughout. Humans demonstrate a more balanced use of both, adapting their responses more contextually.

Overall, LLMs reveal more polarized and rigid strategic tendencies, while humans adjust their behaviors more adaptively, highlighting ongoing challenges in behavioral alignment. Personality-specific patterns were similar across models (see supplementary material).

\subsubsection{Temporal Dynamics of Strategic Behavior Across Human and LLM Dialogues}
\begin{figure}[t!]
    \centering
    \includegraphics[width=\linewidth]{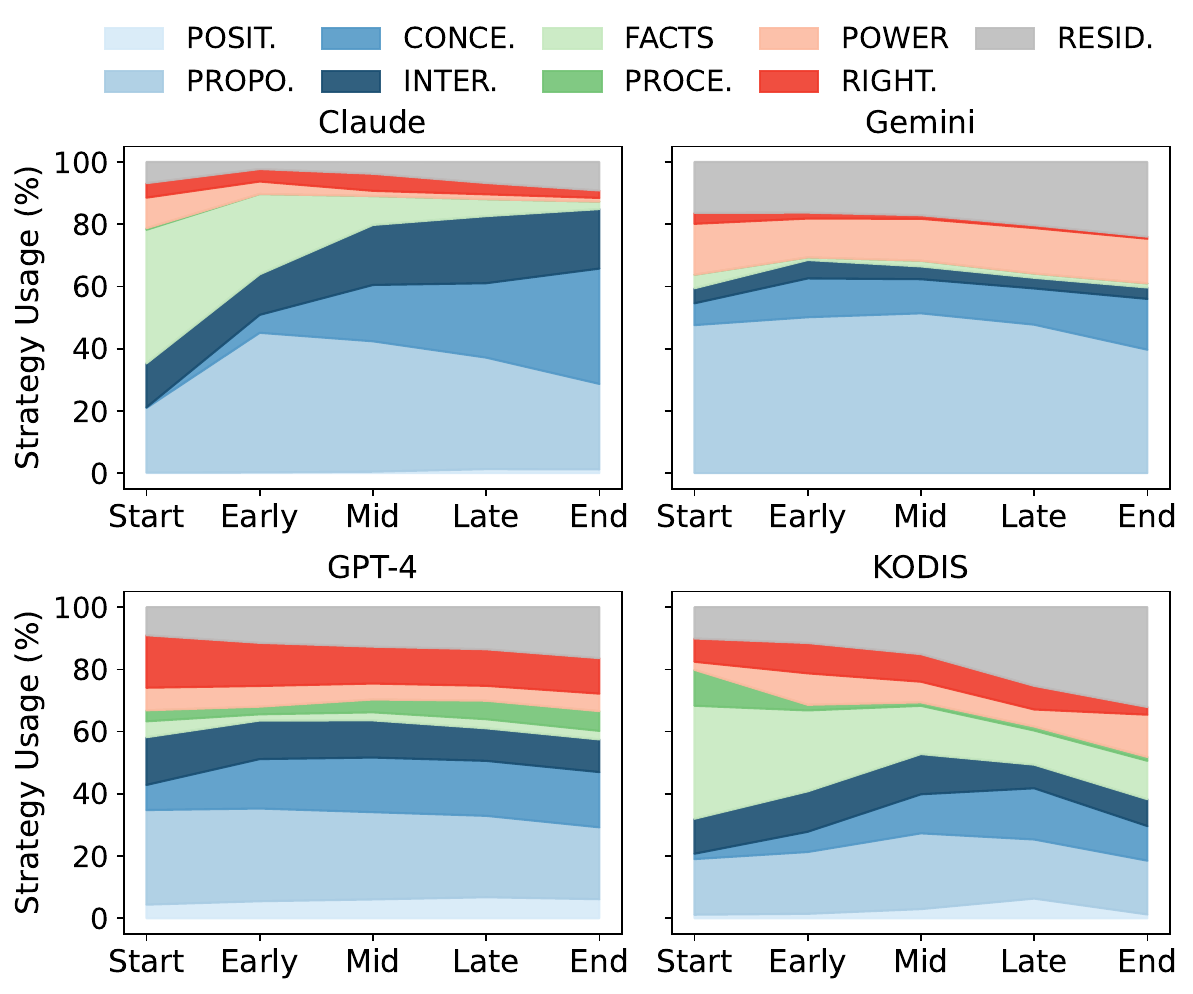}
    \caption{Temporal distribution of the frequency of IRP strategies across dialogue stages in high extraversion cases for human (KODIS) and LLM (L2L) dialog.}
    \label{fig:IRP_temporal_ext}
    \vspace{-2ex}
\end{figure}

Figure~\ref{fig:IRP_temporal_ext} shows IRP strategy distributions over dialogue stages in high-extraversion cases for humans and LLMs. Full results for other traits can be found in supplementary material.
Humans exhibit dynamic progression: \emph{Facts} dominate early stages then decline as \emph{Interest}, \emph{Proposal}, and \emph{Concession} increase, with greater \emph{Residual} use at the end. This reflects structured shifts from factual grounding to relational closure, while \emph{Power} and \emph{Rights} remain minimal.
LLMs show flatter trajectories with limited adaptation. Most begin with dominant \emph{Proposal} use, while \emph{Concession} appears earlier and more persistently than in humans, suggesting premature accommodating behavior. Except for Claude, LLMs rarely use \emph{Facts} early on, bypassing the grounding phase. Claude partially mirrors human dynamics with declining \emph{Facts} and increasing \emph{Interest} and \emph{Concession}. Gemini remains \emph{Proposal}-dominated with high \emph{Power} and \emph{Residual} but little temporal variation. GPT-4 shows balanced use but maintains consistently high \emph{Rights}.
These patterns reveal that while humans adapt strategies temporally, LLMs follow fixed, model-specific paths, highlighting the need for improved temporal flexibility.

\section{Conclusion}
This study provides the first behavior-rich comparison of personality-driven conflict behaviors between humans and LLMs under matched conflict resolution scenarios. Our analyses directly address whether personality-prompted LLMs can replicate human-like behavior in high-stakes interpersonal contexts. While humans flexibly adapt based on both self and partner traits, LLMs show distinct personality-linked patterns that often diverge in how they participate in and resolve conflict, especially in dynamic aspects of resolution rather than static strategy use. These findings highlight key limitations of current prompting approaches and caution against assuming that trait-driven LLMs can reliably proxy human behavior. As LLMs enter socially consequential domains, our results underscore the need for more psychologically grounded and context-sensitive models to ensure safe and responsible deployment.

\section{Limitations and Future Work}
This study offers insights into how personality shapes strategic behavior and outcomes, but limitations remain. First, our LLM-based IRP annotations achieved strong F1 scores and partial human validation, but we did not conduct a full-scale review. Second, we minimized prompt variation to test whether LLMs can inherently represent human personality traits, but even small changes in phrasing, instructions, or ordering may shift model behavior; we did not assess robustness to such variations, limiting the generalizability of our findings. Finally, while BFI traits were informative, other personality constructs (e.g., emotional intelligence, Machiavellianism) may offer additional explanatory power.

Future work should incorporate more naturalistic, multimodal conflict data, strengthen annotation validation, and test generalization across LLMs. Our framework could also be extended to analyze linguistic and emotional patterns and to support LLM alignment with personality–behavior relationships observed in human data.

\section*{Acknowledgements}
Research was sponsored by the Army Research Office under Cooperative Agreement Number W911NF-25-2-0040. The views and conclusions contained in this document are those of the authors and should not be interpreted as representing the official policies, either expressed or implied, of the Army Research Office or the U.S. Government. The U.S. Government is authorized to reproduce and distribute reprints for Government purposes, notwithstanding any copyright notation herein.  Kaleen Shrestha is supported by an NSF CISE Graduate Fellowship CSGrad4US under
Grant No. 2313998 (Award ID G-2A-061). 

\bibliography{aaai2026}
\clearpage
\appendix
\section{LLM Annotation of IRP Strategies}\label{append:LLM_IRP_Annotation}
To obtain IRP strategy labels for the KODIS dataset (examples and definitions can be found in Table \ref{tab:irp-definition-examples}), we leveraged a combination of human evaluation and large language model (LLM)-based annotation. First, we validated the reliability of LLM-generated annotations through extensive human evaluation, followed by full-dataset annotation using GPT-4o \citep{openai2024gpt4ocard} (gpt-4o-2024-08-06, run on 3/19/2025) with the default temperature value of 1. This section details the annotation procedure and evaluation metrics.

\subsection{Inter-Annotator Agreement for Human Evaluation Annotation}
We first had human annotations on a 10\% subset of the KODIS human-to-human conversations (25 conversations). Three annotators (including one of the authors) were trained on nine IRP conflict resolution strategies defined by \citet{brett1998breaking}, omitting \textit{Request for Proposal} following \citet{shaikh2024rehearsal}. Utterances were segmented into subject-verb sequences to account for multiple IRP strategies within a turn.

Annotators initially attempted direct classification, but low inter-annotator agreement led us to shift to an evaluation framework: annotators assessed the correctness of GPT-4o predictions as binary correct/incorrect labels. Due to the imbalance in label distribution and prevalence of majority labels, Fleiss' Kappa was not representative. We therefore used A-Kappa \citep{gautam2014kappa}, which adjusts for label imbalance.

Table~\ref{tab:a-kappa-irp} presents the A-Kappa scores for each IRP strategy based on human evaluation. All IRP categories achieved an A-Kappa score of at least 0.80, indicating strong inter-annotator agreement on the correctness of LLM annotations.

Figure~\ref{fig:irp_prompt_snippet} displays the snippet of the prompt used for the IRP annotations.
\begin{figure}[h!]
\centering
\begin{myexampleblock}{IRP Annotation Prompt Snippet}
\small{
\# \textbf{IRP Strategy Definitions and Examples} \\
\ [Cooperative Strategies]\\\\
\ INTERESTS: Reference to the wants, needs, or concerns of one or both parties. This may include questions about why the negotiator wants or feels the way they do. This does not include anything about wanting a deal (apology, refund, removing negative review) without a reason.\\\\
\ Example: “I understand that you've been really busy lately.” \\
\ Non-example: "I don't understand."\\
\ ...\\\\
\# \textbf{Annotation Instructions} \\
You need to annotate the following conversation at the utterance level, identifying which strategy from the IRP framework aligns with each sentence... \\ \\
}\end{myexampleblock}
\caption{IRP Annotation Prompt for GPT-4o}
\label{fig:irp_prompt_snippet}
\end{figure}

\subsection{IRP Annotation Evaluation}
After validating the annotation quality through human evaluation, we used GPT-4o to annotate the full KODIS dataset. An overview of our prompt can be found in Figure \ref{fig:irp_prompt_snippet}. Predictions judged incorrect during human evaluation were further deliberated, while correct predictions were retained as gold labels.

The final LLM-based annotation achieved an overall accuracy of 81\%, a macro-average F1 score of 79\%, and a weighted-average F1 score of 81\% on the held-out evaluation set. This performance is comparable to existing IRP classification work by \citet{shaikh2024rehearsal}, which reported an average accuracy of 82\% (with the lowest class accuracy of 66\%).

Table~\ref{tab:f1-score-llm-irp} presents few-shot classification F1 scores across the IRP strategies. Among them, the \textit{Positive Expectations} category achieved the lowest F1 score of 0.69, which remains comparable or slightly better relative to prior studies. The final prompt used for the GPT-4o annotations is included later in the supplementary material.

\section{Distribution of IRP Strategies}

Figure~\ref{fig:all_strategies_model_comparison} shows
the distribution of IRP strategy frequencies across high-trait personalities in LLMs and humans. Clear behavioral differences between LLMs and humans are observed for each strategy, while differences across personality traits within each group are less pronounced.


\section{Regression}
We present full linear regression results for personality and conflict resolution behavior measures. 

\subsection{Regression Models}\label{append:regrssion_formula}
The following regression models were used to examine the relationship between personality traits and negotiation behaviors:
\begin{equation}
\text{DV}_k = \beta_0 + \sum_{j=1}^{5} \beta_j \text{SELF}_j + \sum_{j=1}^{5} \gamma_j \text{PARTNER}_j + \beta_{11} \text{Position} 
\end{equation}
where  $\text{DV}_k$ denotes the k-th dependent variable among the set of negotiation metrics. $\text{SELF}_j$ and $\text{PARTNER}_j$ represent the j-th BFI trait (extraversion (EXT), conscientiousness (CON), agreeableness (AGR), neuroticism (NEU), and openness to experience (OPE)) of the player and their partner, respectively. 

\subsection{Details of Dependent Variables}\label{append:DV_details}

Table~\ref{tab:details_DVs} provides a summary of dependent variables used in the regression analyses, categorized by Outcome from strategic choices and Strategic Behavior.


\begin{table}[h]
\centering
\renewcommand{\arraystretch}{1}
\resizebox{1\linewidth}{!}{%
\begin{tabular}{llll}
\hline
\multicolumn{1}{c}{\begin{tabular}[c]{@{}c@{}}Dependent \\ Variable (DV)\end{tabular}} & \multicolumn{1}{c}{Type} & \multicolumn{1}{c}{Value} & \multicolumn{1}{c}{\begin{tabular}[c]{@{}c@{}}Regression \\ Model\end{tabular}} \\ \hline
\multicolumn{4}{c}{\cellcolor[HTML]{EFEFEF}Outcomes DVs}\\ \hline
Score  & Continuous& 0-100 & OLS Reg.  \\
Walk-Away   & Binary& 0 or 1& Logistic Reg.  \\
Accept & Binary& 0 or 1& Logistic Reg.  \\ \hline
\multicolumn{4}{c}{\cellcolor[HTML]{EFEFEF}Strategic Behavior DVs}   \\ \hline
IRP Ratio   & Continuous& 0-100 & OLS Reg.  \\
IRP Reciprocity  & Continuous& 0-100 & OLS Reg.  \\ 
Escalation and De-escalation Ratio   & Continuous& 0-100 & OLS Reg.  \\\hline
\end{tabular}}
\caption{Dependent Variables and Regression Models for Regression Analyses}
\label{tab:details_DVs}
\end{table}

\subsection{Personality and Participation and Resolution Behavior}

Table \ref{tab:regression_full_results_participation_resolution} contains the full regression results for L2L and KODIS datasets for the conflict resolution behavior metrics.

\subsection{Personality and Strategic Conflict Resolution Behavior}\label{append:IRP_detailed_interpret}

\subsubsection{Regression Results for Individual IRP Ratio}

This section presents detailed regression results for personality effects on the usage (i.e., ratio) of individual IRP strategies in the KODIS dataset.

\noindent \textbf{[\emph{Cooperative} Strategies]}
\begin{itemize}
\item \textbf{Positive Expectations}: Self-neuroticism was negatively associated with the use of \emph{Positive Expectations} (B=-1.3, p=.02), whereas self-openness to experience was positively associated (B=1.4, p=.02). These findings suggest that individuals higher in neuroticism may be less likely to make optimistic statements, whereas those higher in openness may be more inclined to recognize shared goals and express optimism.
\item \textbf{Proposals}: Partner neuroticism was negatively associated with \emph{Proposal} usage (B=-2.3, p=.02). Participants interacting with more neurotic partners may have been less willing to reciprocate with assertive \emph{Proposal} to avoid triggering negative emotional reactions.
\item \textbf{Concessions}: Partner neuroticism positively predicted concession behavior (B=1.8, p=.01), indicating that participants were more likely to concede when faced with more emotionally reactive partners. Additionally, partner openness to experience was positively associated with \emph{Concessions} (B=1.8, p=.04), suggesting that creativity and receptiveness facilitated greater acceptance and accommodation.
\end{itemize}

\noindent \textbf{[\emph{Competitive} Strategies]}
\begin{itemize}
\item \textbf{Rights}: Partner conscientiousness was negatively associated with \emph{Rights} strategy usage (B=-1.8, p=.05). Participants may have been less inclined to reciprocate \emph{Rights}-based moves when interacting with highly conscientious partners, whose arguments may have been more thorough and difficult to refute.
\end{itemize}

See Table~\ref{tab:irp_full_regressions_ratio} for the complete regression results at the individual strategy level.

\begin{figure}[th]
\centering
\begin{myexampleblock}{LLM Prompt Snippet (Buyer)}
\small{
\# \textbf{Personality} \\
\{Personality Adjectives\} \\ \\
\# \textbf{Story} \\\textbf{}
You purchased a Kobe Bryant championship jersey for your terminally ill nephew for \$75 from a site ... \\ \\
\# \textbf{Instructions} \\
You (Buyer) are now chatting with this Seller; respond to the dialog history through text messages.\\ 

\#\# \textbf{Issues to resolve} \\
1. Refund (REF), 2. Seller Negative Review (SNR),  \\ 3. Buyer Negative Review (BNR),  4. Seller Apology (SAP),  5. Buyer Apology (BAP) \\ \\
\#\# \textbf{Issues Importance} \\
This shows how important each issue is to you. The final outcome score is computed as the inner product of each agreed-upon value and your assigned importance weights. 
... \\\\
\#\# \textbf{Strategy \& Behavior Rules} \\
- You are encouraged to explore alternative solutions that better reflect your issue priorities
- Always offer a clear trade-off (ex., at least one compromise or concession) if rejecting an offer. \\
- If your partner clearly refuses one issue (e.g., “I cannot apologize”), shift your strategy ... \\

\#\# \textbf{Required Multi-Issue Engagement} \\
You must negotiate across at least 3 issues. Do not focus only on a single issue. \\

\#\# \textbf{Sample Dialogue Examples (Reference Only).} \\
\{Sample Dialogs\}
...
}\end{myexampleblock}
\caption{LLM Prompt Snippet as a Buyer}
\label{fig:prompt_snippet}
\end{figure}

\begin{table*}[t!]
\centering
\small
\renewcommand{\arraystretch}{1}
\resizebox{0.9\linewidth}{!}{%
\begin{tabular}{@{}cl@{}}
\toprule
\textbf{Speaker} & \multicolumn{1}{c}{\textbf{Conversation}}  \\ \midrule
\rowcolor[HTML]{EFEFEF} 
Buyer& I need you to remove your false review about me and apologize for misleading others about the jersey.  \\
Seller & \begin{tabular}[c]{@{}l@{}}I won't remove my truthful review when you're the one who lied calling me a "LIAR AND A CROOK" \\ - the listing clearly stated it wasn't for a specific player.\end{tabular}\\
\rowcolor[HTML]{EFEFEF} 
Buyer& \begin{tabular}[c]{@{}l@{}}Your listing specifically mentioned Kobe Bryant and you changed it later \\ - I'll reconsider my review if you remove yours and offer a public apology.\end{tabular}  \\
Seller & Never mentioned Kobe in my listing - check your screenshots if you have any proof.   \\
\rowcolor[HTML]{EFEFEF} 
Buyer& \begin{tabular}[c]{@{}l@{}}I saw the Kobe reference before you changed it, but I'm willing to meet halfway \\ - remove your harmful review about me and apologize, and we can discuss my review separately.\end{tabular} \\
Seller & I'll consider removing my review if you take down your false review and apologize for calling me a liar and crook. \\
\rowcolor[HTML]{EFEFEF} 
Buyer& \begin{tabular}[c]{@{}l@{}}I need your apology more than anything since you've damaged my reputation, \\ but I'm willing to adjust my review if you remove yours first.\end{tabular} \\
Seller & \begin{tabular}[c]{@{}l@{}}Look, I care about receiving an apology from you \\ - your false claims damaged my business reputation far more than my review hurt you.\end{tabular} \\
\rowcolor[HTML]{EFEFEF} 
Buyer& \begin{tabular}[c]{@{}l@{}}I understand reputations matter to both of us \\ - how about we both remove our reviews and you apologize for the misrepresentation about the jersey?\end{tabular}\\
Seller & \begin{tabular}[c]{@{}l@{}}I'll apologize if you remove your review and apologize for falsely calling me a liar \\ - the jersey was never misrepresented.\end{tabular}   \\
\rowcolor[HTML]{EFEFEF} 
Buyer& \begin{tabular}[c]{@{}l@{}}Your apology is most important to me - if you sincerely apologize and remove your review, \\ I'll consider adjusting my review without asking for a refund.\end{tabular}\\
Seller & \begin{tabular}[c]{@{}l@{}}Since you're dropping the refund demand, I'll apologize but I need you to remove your review completely \\ - that's my priority.\end{tabular} \\
\rowcolor[HTML]{EFEFEF} 
Buyer& \begin{tabular}[c]{@{}l@{}}I'll agree to remove my review if you apologize sincerely and remove your false review about me \\ - that seems fair for both of us.\end{tabular}   \\
Seller & SUBMISSION: \{"REF": "None", "SNR": "remove", "BNR": "remove", "SAP": "apologize", "BAP": "not apologize"\}  \\
\rowcolor[HTML]{EFEFEF} 
Buyer& ACCEPT-DEAL  \\ \bottomrule
\end{tabular}
}
\caption{Example L2L simulation dialogue generated by Claude-Sonnet-3.7. The negotiation issue abbreviations are: REF = refund, SNR/BNR = Seller/Buyer negative review, SAP/BAP = Seller/Buyer apology.}
\label{tab:L2L_dialogue_example}
\end{table*}

\subsubsection{Regression Results for Individual IRP Strategy Reciprocity}

This section presents detailed regression results for personality effects on reciprocity at the level of individual IRP strategies in the KOIDS dataset.

\noindent \textbf{[\emph{Cooperative} Strategies]}
\begin{itemize}
\item \textbf{Proposals}: \emph{Proposal} reciprocity was negatively associated with partner neuroticism (B=-5.1, p=.01), suggesting that participants may have been less likely to reciprocate \emph{P}s when interacting with more emotionally reactive partners.
\item \textbf{Concessions}: \emph{Concession} reciprocity was negatively associated with self openness to experience (B=-6.6, p=.04), possibly reflecting a reduced tendency among more open individuals to conform to normative reciprocal behaviors.
\end{itemize}

\noindent\textbf{[\emph{Competitive} Strategies]}
\begin{itemize}
\item \textbf{Rights}: \emph{Rights} strategy reciprocity was negatively associated with partner conscientiousness (B=-9.3, p=.002), indicating that participants were less willing to reciprocate \emph{Rights}-based moves when faced with highly conscientious partners.
\end{itemize}

Please refer to Tables ~\ref{tab:irp_full_regressions_reciprocity} for the complete regression results at the individual strategy level.

\section{LLM-to-LLM (L2L) Simulation} \label{sec:appendix-a2a}

\subsection{Hyperparameters of LLMs} 
We used default hyperparameters for all LLMs:\\
– GPT-4o mini: temperature = 1.0, top-1\\
– Claude-3.7-Sonnet: temperature = 1.0, top-p = 0.99\\
– Gemini-Flash-2.0: temperature = 1.0, top-p = 0.95

\subsection{LLM prompt \& Dialogue example} 
The Figure~\ref{fig:prompt_snippet} displays the LLM prompt snippet used for L2L simulation as Buyer. An example of an L2L dialog (generated by Claude Sonnet 3.7) can be found in Table \ref{tab:L2L_dialogue_example}. This dialog, like all L2L data, are generated with LLMs prompted with the KODIS scenario and matched personality profile sampled from the personality profile distribution found in KODIS. 

\subsection{Personality Matching} 
\begin{figure}[h]
\centering
\includegraphics[width=1\linewidth]{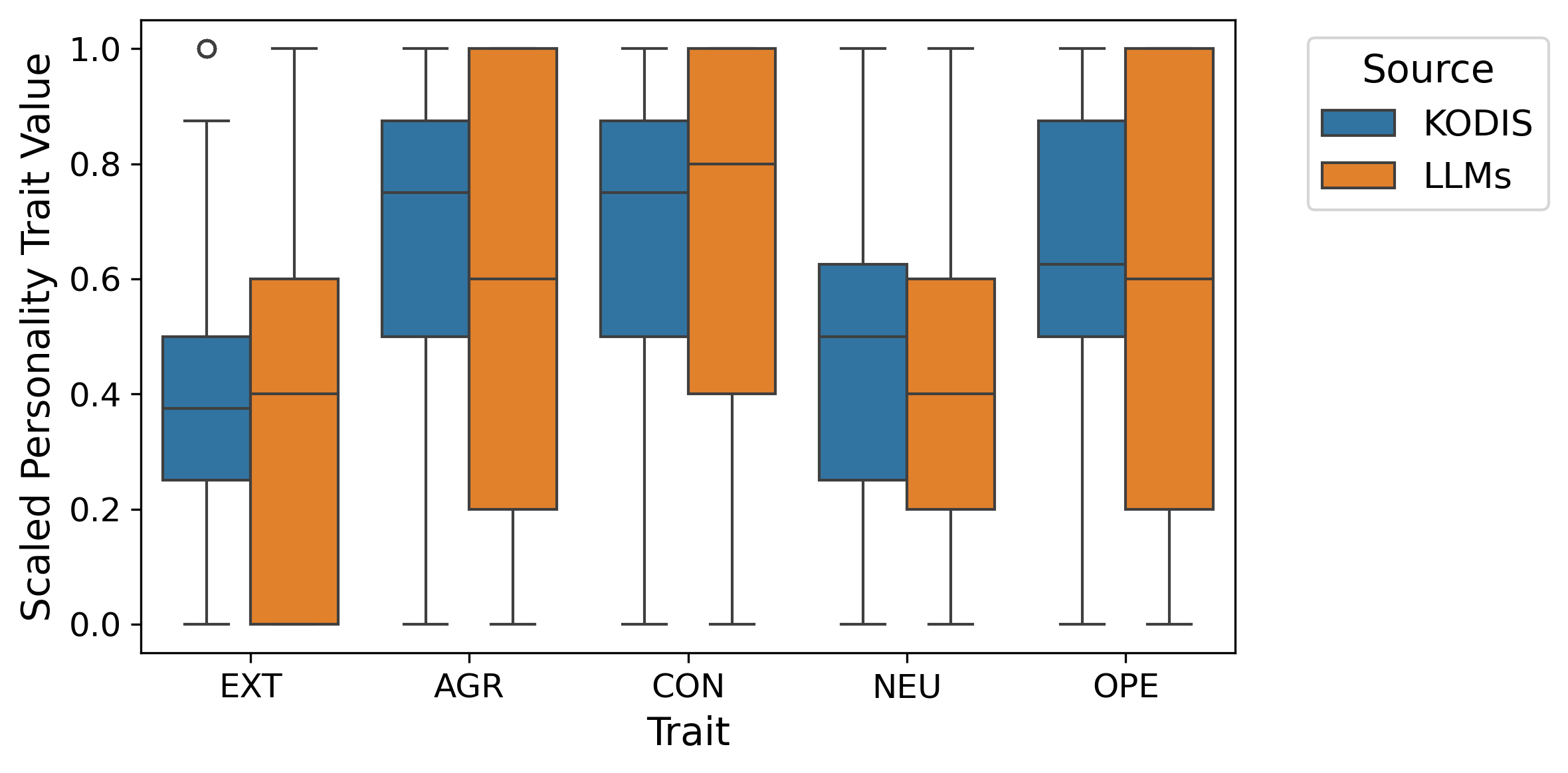}
\caption{Distribution of scaled BFI personality trait values across human participants (KODIS) and LLMs.}
\label{fig:personality_matched_dist}
\end{figure}

To enable fair comparisons between humans and LLMs in downstream behavioral analysis, we generate personality profiles for LLMs through weighted sampling based on the empirical distribution of human traits. As shown in Figure~\ref{fig:personality_matched_dist}, the resulting trait-wise personality distributions for LLMs (GPT-4o-mini) closely mirror those of human participants. This alignment ensures that the LLMs involved in the simulation exhibit human-like personality profiles, thereby supporting more meaningful and controlled human–LLM comparisons.

\subsection{Data}
Below are the details for all L2L datasets:
\subsubsection{Claude Sonnet 3.7 L2L Dataset}
English dyadic conversation text dataset, generated by two Claude Sonnet 3.7 LLMs. 250 dialogs, each with at most 25 rounds. Topic is based on the KODIS dataset, where a Buyer and Seller are discussing a dispute over a Kobe Bryant basketball jersey, and are attempting to come to an agreement on how to proceed (see Methodology section). 
\subsubsection{Gemini 2.0 Flash L2L Dataset} 
English dyadic conversation text dataset, generated by two Claude Sonnet 3.7 LLMs. 250 dialogs, each with at most 25 rounds. Topic is based on the KODIS dataset, where a Buyer and Seller are discussing a dispute over a Kobe Bryant basketball jersey, and are attempting to come to an agreement on how to proceed (see Methodology section). \subsubsection{GPT-4o mini L2L Dataset}
English dyadic conversation text dataset, generated by two Claude Sonnet 3.7 LLMs. 500 dialogs, each with at most 25 rounds. Topic is based on the KODIS dataset, where a Buyer and Seller are discussing a dispute over a Kobe Bryant basketball jersey, and are attempting to come to an agreement on how to proceed (see Methodology section). 

\section{Results}

\noindent \textbf{Role-Contingent Personality Effects} We want to assess whether personality traits influence negotiation behavior differently based on role (Buyer (\textit{B}) vs. 
Seller(\textit{S})). We replaced the effect-coded position variable with a dummy-coded version (0=\textit{B}, 1=\textit{S}) and included interaction terms between position and each self-trait. This allowed us to examine role-contingent trait effects using both regression coefficients and simple effect analyses derived from linear combinations (Table \ref{tab:regression_dummy_simple_effect}).

For the offer acceptance decision, role-dependent effects emerged across models but varied by trait. In both GPT-4o-mini and KODIS, higher agreeableness significantly reduced acceptance in the Buyer role only (GPT: B=–0.14, p = .043; KODIS: B=–0.39, p =.049), showing selective sensitivity to role. In contrast, Gemini 2.0 Flash showed role-contingent effects for neuroticism rather than agreeableness—acceptance increased with neuroticism overall (B=0.30, p=.014) but decreased in the Buyer role (B = –0.48, p=.006). These patterns highlight that while GPT-4o-mini and humans modulated acceptance based on agreeableness and role, Gemini’s behavior was more strongly shaped by role-dependent neuroticism.

This alignment in role-contingent behavior suggests that LLMs can partially mirror human decision patterns, particularly in offer acceptance. However, such alignment does not extend to performance outcomes. For score, only GPT-4o showed a significant role-dependent effect of agreeableness, with lower scores in both Buyer (B = –2.62, p=.001) and Seller roles (B = –3.32, p$<$.001). Gemini 2.0 Flash showed no significant effect in the Seller role, and no personality traits predicted human scores. This indicates that, unlike humans, LLMs apply trait-based behaviors more uniformly across roles, with limited sensitivity to context.

Overall, while certain personality-contingent behaviors in LLMs mirror human patterns, this alignment appears limited to a specific domain. These findings suggest that current LLM architectures may offer partial behavioral alignment with human-like role sensitivity, but still show limited adaptive flexibility and responsiveness to broader social context.

\begin{table}[h!]
\centering
\small
\renewcommand{\arraystretch}{1}
\resizebox{1\linewidth}{!}{%
\begin{tabular}{@{}llccc@{}}
\toprule
\multicolumn{1}{c}{}& \multicolumn{1}{c}{}& \multicolumn{3}{c}{Beta Coefficient (p-value)}\\ \cmidrule(l){3-5} 
\multicolumn{1}{c}{\multirow{-2}{*}{DVs}}& \multicolumn{1}{c}{\multirow{-2}{*}{Significant IVs}} & GPT-4o mini   & Gemini 2.0 Flash & KODIS \\ \midrule
ACCEPT& S-AGR& -0.14* & & -0.39*\\
& S-AGR × POS & 0.26** & & 0.54* \\
& S-NEU& & 0.30*&\\
& S-NEU × POS & & -0.48**&\\ \cmidrule(l){2-5} 
\multicolumn{1}{c}{}& \cellcolor[HTML]{ECF4FF}Buyer@POS=0 & \cellcolor[HTML]{ECF4FF}-0.14* & \cellcolor[HTML]{ECF4FF}0.30*   & \cellcolor[HTML]{ECF4FF}-0.39* \\
\multicolumn{1}{c}{\multirow{-2}{*}{\begin{tabular}[c]{@{}c@{}}(Simple \\ Effect)\end{tabular}}} & \cellcolor[HTML]{ECF4FF}Seller@POS=1& \cellcolor[HTML]{ECF4FF}0.10& \cellcolor[HTML]{ECF4FF}\textbf{0.29*} & \cellcolor[HTML]{ECF4FF}0.05   \\ \midrule
SCORE & S-AGR& -2.62***& &\\
& S-AGR × POS & -3.40**& &\\
& S-NEU& & -3.46**&\\
& S-NEU × POS & & 4.84** &\\ \cmidrule(l){2-5} 
\multicolumn{1}{c}{}& \cellcolor[HTML]{ECF4FF}Buyer@POS=0 & \cellcolor[HTML]{ECF4FF}-2.62***   & \cellcolor[HTML]{ECF4FF}-3.46** & \cellcolor[HTML]{ECF4FF}\\
\multicolumn{1}{c}{\multirow{-2}{*}{\begin{tabular}[c]{@{}c@{}}(Simple \\ Effect)\end{tabular}}} & \cellcolor[HTML]{ECF4FF}Seller@POS=1& \cellcolor[HTML]{ECF4FF}\textbf{-3.32***} & \cellcolor[HTML]{ECF4FF}1.39& \cellcolor[HTML]{ECF4FF}\\ \bottomrule
\end{tabular}
}
\caption{Summary of significant personality effects and their interactions with role in L2L and KODIS. Simple effects computed using reverse-coded position (0 = Seller).}
\label{tab:regression_dummy_simple_effect}
\end{table}

\subsubsection{Full Regression Results.} Tables~\ref{tab:regression_full_results_participation_resolution}, \ref{tab:irp_full_regressions_ratio}, \ref{tab:irp_full_regressions_reciprocity}, and \ref{tab:escalation} present the full regression results for key outcome and strategy variables used in the LLMs and human comparisons, including strategic outcome variable (e.g., score, accept), IRP ratios, reciprocity, and (de)escalation behaviors.

\subsubsection{High-Trait Personality Cases.} 
\begin{figure*}[h]
\centering
\includegraphics[width=1\linewidth]{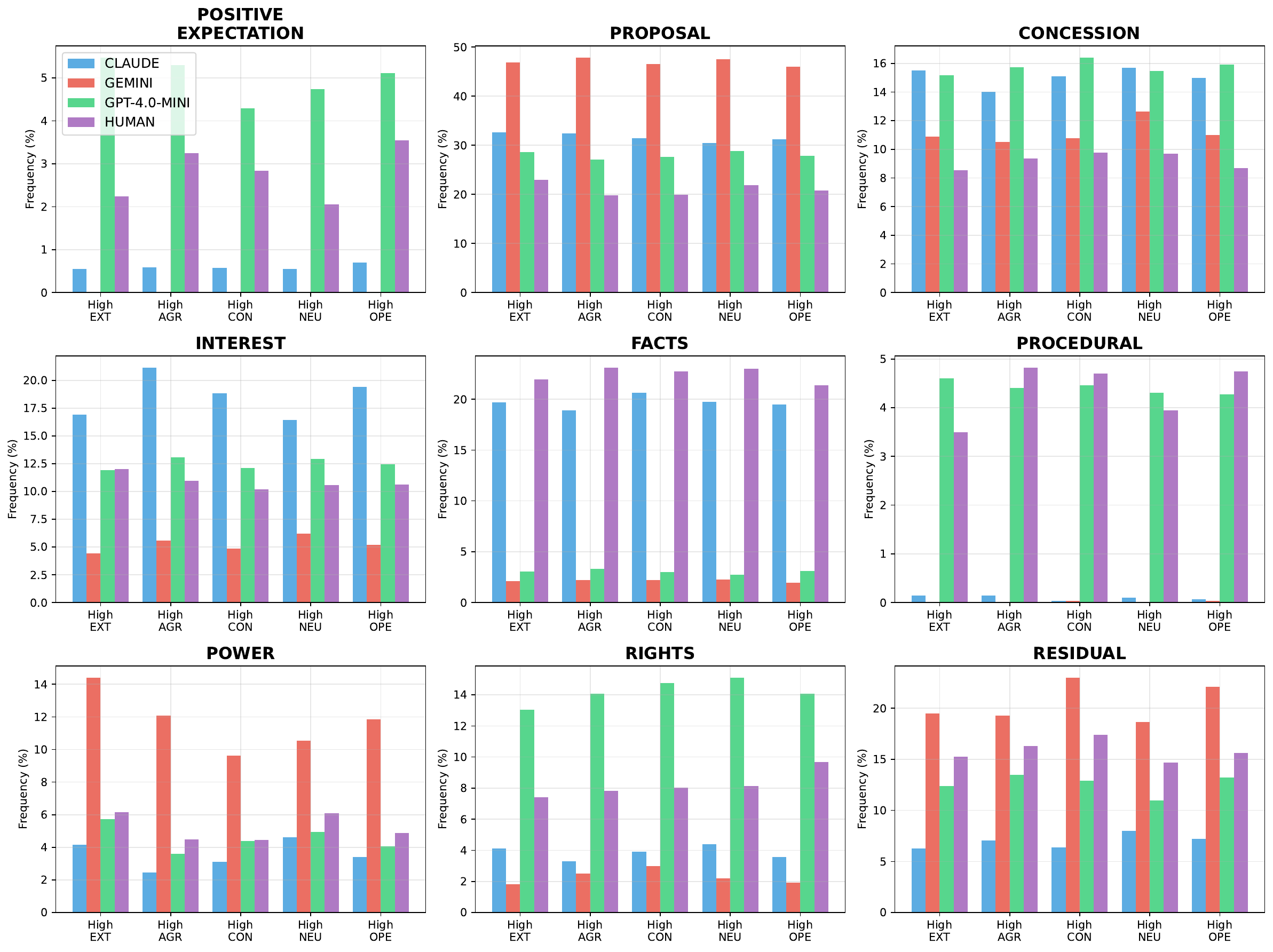}
\caption{Distribution of IRP frequencies across high-trait personality cases by strategy in the LLMs and humans (KODIS). High-trait cases refer to trait levels $\geq$ "moderate" in L2L and $>$ 3.5 in KODIS.}
\label{fig:all_strategies_model_comparison}
\end{figure*}

\begin{figure*}[ht]
\centering
\includegraphics[width=1\linewidth]{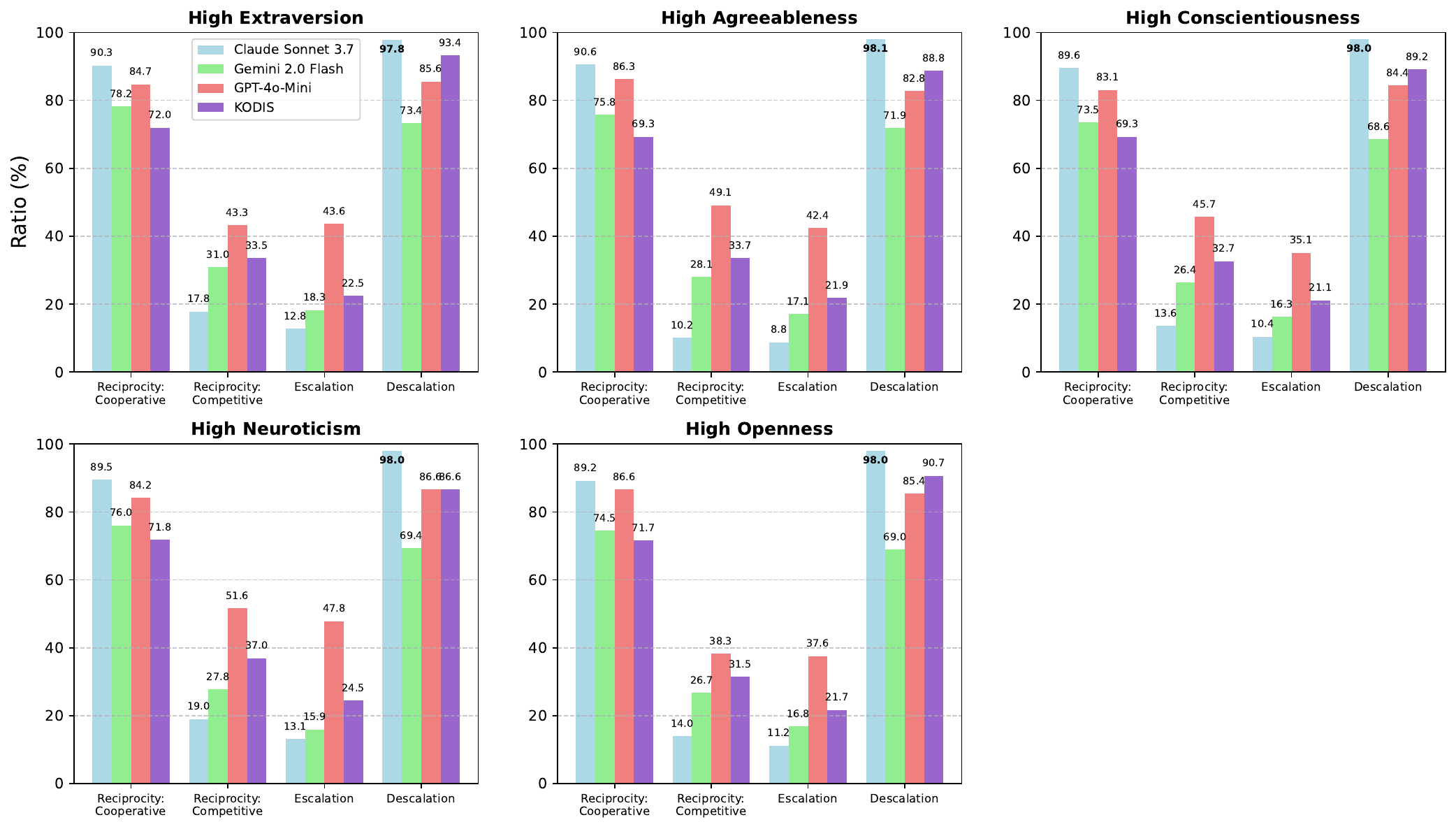}
\caption{Distribution of IRP strategy reciprocity and (de)escalation frequencies across high-level personality trait cases in the LLMs’ simulations (L2L) and human conversations (KODIS) datasets. High-trait cases refer to trait levels $\geq$ "moderate" in L2L and $>$ 3.5 in KODIS.
}
\label{fig:irp_recip_dist_personality_full}
\end{figure*}

\begin{figure*}[h]
\centering
\includegraphics[width=1\linewidth]{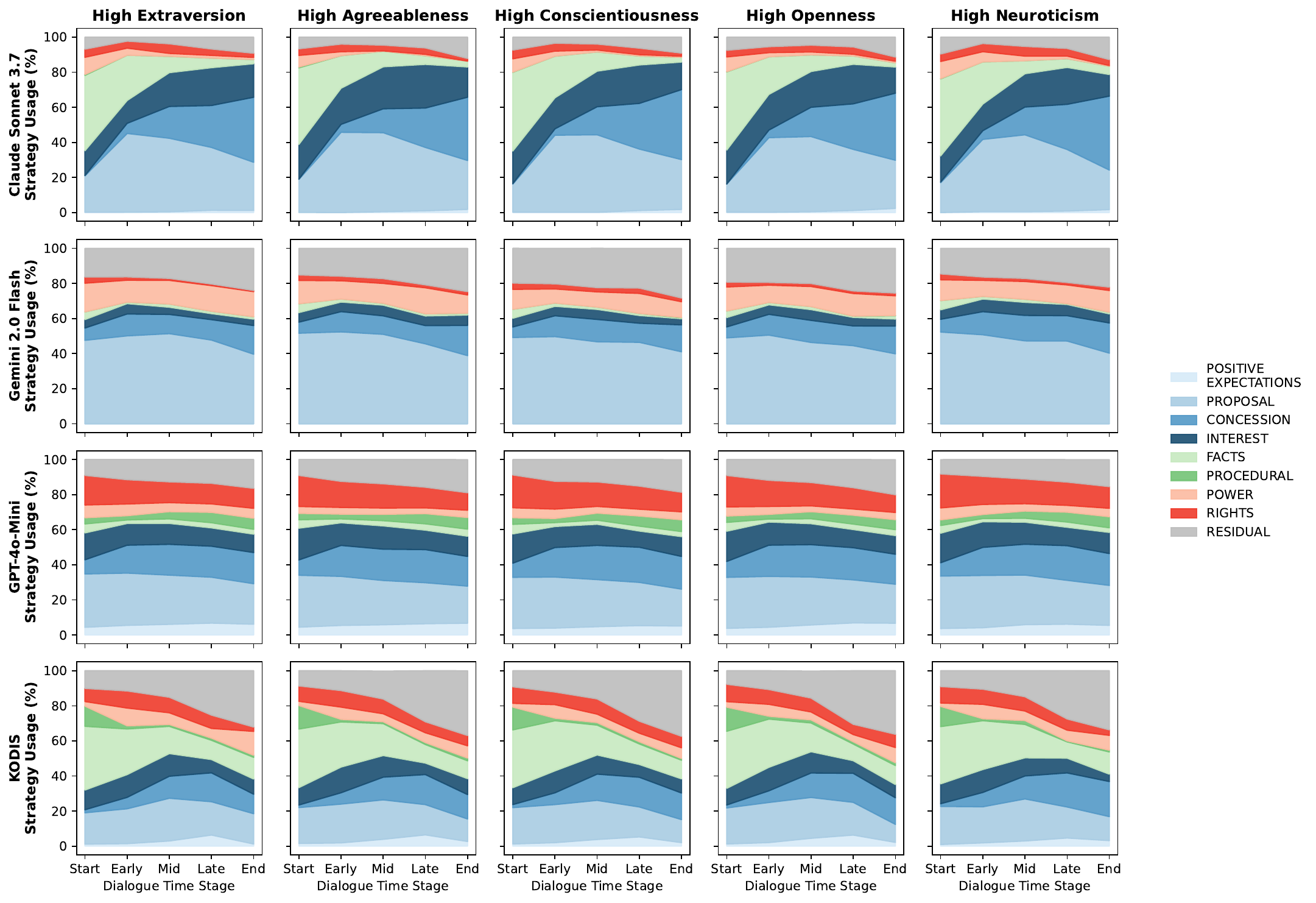}
\caption{Temporal distribution of IRP strategy usage across high-trait personality cases in the LLMs' simulations (L2L) and humans (KODIS). High-trait cases refer to trait levels $\geq$ "moderate" in L2L and $>$ 3.5 in KODIS.}
\label{fig:irp_temporal_dist_personality}
\end{figure*}

Figures~\ref{fig:all_strategies_model_comparison}, \ref{fig:irp_recip_dist_personality_full}, and \ref{fig:irp_temporal_dist_personality} show the IRP ratio, RP strategy reciprocity and (de)escalation frequencies, and the temporal distribution of IRP strategy usage for high-trait cases across the five personality dimensions.

\clearpage


\begin{table*}[]
\resizebox{\textwidth}{!}{%
\begin{tabular}{llll}
\hline
\multicolumn{1}{c}{IRP Category}& \multicolumn{1}{c}{IRP Strategy} & \multicolumn{1}{c}{Definition} & \multicolumn{1}{c}{Example} \\ \hline
\multirow{4}{*}{\textit{Cooperative}} & \textit{Proposal}   & \begin{tabular}[c]{@{}l@{}}Concrete solution ideas that may \\ resolve the conflict.\end{tabular}   & \begin{tabular}[c]{@{}l@{}}The best offer I can give \\ you is a partial refund, how does that sound?\end{tabular}   \\ \cline{2-4} 
& \textit{Concession} & \begin{tabular}[c]{@{}l@{}}Change in initial view in \\ response to \emph{Proposal}.\end{tabular}& \begin{tabular}[c]{@{}l@{}}Ok fine, I will give you \\ a refund instead.\end{tabular}\\ \cline{2-4} 
& \textit{Interests}  & \begin{tabular}[c]{@{}l@{}}Referencing need, wants, \\ and concerns of either side.\end{tabular}& \begin{tabular}[c]{@{}l@{}}I understand you want \\ this refund because of your nephew\end{tabular} \\ \cline{2-4} 
& \textit{Positive Expectations}   & \begin{tabular}[c]{@{}l@{}}Expressing positive outlook \\ by recognizing common goals or similarities\end{tabular}& \begin{tabular}[c]{@{}l@{}}You and I both want to \\ conclude this conversation well.\end{tabular}\\ \cline{2-4} 
\multirow{2}{*}{\textit{Neutral}} & \textit{Facts}& \begin{tabular}[c]{@{}l@{}}Statement clarifying or \\ requesting information about the situation.\end{tabular}   & \begin{tabular}[c]{@{}l@{}}The product you bought \\ was not from my website.\end{tabular} \\ \cline{2-4} 
& \textit{Procedural} & \begin{tabular}[c]{@{}l@{}}Statements related to process or \\ rules of the negotiation process, \\ or introductory messages.\end{tabular} & \begin{tabular}[c]{@{}l@{}}Hello, can we please \\ talk about this issue?\end{tabular} \\ \cline{2-4} 
\multirow{2}{*}{\textit{Competitive}} & \textit{Power}& \begin{tabular}[c]{@{}l@{}}Statements that include threats, \\ accusations to try to coerce a resolution.\end{tabular}  & \begin{tabular}[c]{@{}l@{}}You are a liar, I will \\ write more negative things about you!\end{tabular} \\ \cline{2-4} 
& \textit{Rights} & \begin{tabular}[c]{@{}l@{}}Statements that reference \\ norms, rules, or fairness.\end{tabular} & \begin{tabular}[c]{@{}l@{}}According to the policy, \\ I cannot give you a refund\end{tabular}   \\ \cline{2-4} 
\textit{Residual}  & \textit{Residual}   & \begin{tabular}[c]{@{}l@{}}If an utterance does not fit \\ into any of the above.\end{tabular}& \begin{tabular}[c]{@{}l@{}}Mostly consists of apologies \\ ("I am sorry"), affirmations \\ ("Ok I will"), and thank you's.\end{tabular} \\ \hline
\end{tabular}%
}
\caption{IRP strategy definitions and examples.}
\label{tab:irp-definition-examples}
\end{table*}


\begin{table*}[h]
\centering
\begin{tabular}{cc}\hline
   IRP Category & A-Kappa (\citet{gautam2014kappa})\\\hline
   \emph{Residual} & 0.86\\
   \emph{Concession} & 0.84\\
   \emph{Positive Expectations} & 0.86\\
   \emph{Rights} & 0.87\\
   \emph{Power} & 0.89\\
   \emph{Procedural} & 0.87\\
   \emph{Facts} & 0.81\\
   \emph{Interest} & 0.82 \\
   \emph{Proposal} & 0.85\\\hline
\end{tabular}
\caption{A-Kappa scores for each IRP strategy category}
\label{tab:a-kappa-irp}
\end{table*}


\begin{table*}[h]
\centering
\begin{tabular}{l|ll}
\hline
Category& Strategy   & F1 Score \\ \hline
\emph{Cooperative} && 0.77 \\ \hline
& \emph{Concession} & 0.79 \\
& \emph{Interests}  & 0.75 \\
& \emph{Positive Expectations}  & 0.69 \\
& \emph{Proposal}   & 0.83 \\ \hline
\emph{Competitive} && 0.81 \\ \hline
& \emph{Power}& 0.78 \\
& \emph{Rights} & 0.83 \\ \hline
\emph{Neutral} && 0.84 \\ \hline
& \emph{Facts}& 0.86 \\
& \emph{Procedural} & 0.81 \\ \hline
\emph{Residual}&& 0.75 \\ \hline
& \emph{Residual}   & 0.75 \\ \hline
& Avg. F1& 0.79 \\
& Weighted avg. F1 & 0.81 \\
& Accuracy   & 0.81 \\ \hline
\end{tabular}
\caption{Few-shot classification F1 score perfomance of our final prompted GPT-4o model. The model was evaluated on 10\% of the KODIS human vs. human dataset (25 conversations).}
\label{tab:f1-score-llm-irp}
\end{table*}

\begin{table*}[thb!]
\centering
\renewcommand{\arraystretch}{1}
\resizebox{1\linewidth}{!}{%
\begin{tabular}{@{}llllll|llllll@{}}
\cmidrule(r){1-5} \cmidrule(l){8-12}
\multicolumn{1}{c}{\multirow{2}{*}{Dataset}} & \multicolumn{1}{c}{\multirow{2}{*}{\begin{tabular}[c]{@{}c@{}}Independent\\ Variables\end{tabular}}} & \multicolumn{3}{c}{Dependent Variables} &   &  & \multicolumn{1}{c}{\multirow{2}{*}{Dataset}} & \multicolumn{1}{c}{\multirow{2}{*}{\begin{tabular}[c]{@{}c@{}}Independent\\ Variables\end{tabular}}} & \multicolumn{3}{c}{Dependent Variables}  \\ \cmidrule(lr){3-5} \cmidrule(l){10-12} 
\multicolumn{1}{c}{} & \multicolumn{1}{c}{} & \multicolumn{1}{c}{\textit{SCORE}} & \multicolumn{1}{c}{\textit{ACCEPT}} & \multicolumn{1}{c}{\textit{NOT WALKAWAY}} & \multicolumn{1}{c|}{} & \multicolumn{1}{c}{} & \multicolumn{1}{c}{} & \multicolumn{1}{c}{} & \multicolumn{1}{c}{\textit{SCORE}} & \multicolumn{1}{c}{\textit{ACCEPT}} & \multicolumn{1}{c}{\textit{NOT WALKAWAY}} \\ \cmidrule(r){1-5} \cmidrule(l){8-12} 
\multirow{12}{*}{Gemini 2.0 flash}   & CONST& 39.98*** (0.000)   & 0.00 (1.000)   & 0.00 (1.000)  &   &  & \multirow{12}{*}{GPT-4o-mini}& CONST& 36.82*** (0.000)   & 0.00 (1.000)& -0.00 (1.000) \\
 & SELF\_EXT& 0.20 (0.839)   & 0.12 (0.200)   & -0.00 (0.974)  &   &  &  & SELF\_EXT& 1.67** (0.005) & 0.01 (0.806)& 0.07 (0.499) \\
 & SELF\_AGR& -4.48*** (0.000)   & 0.04 (0.582)   & 0.14 (0.060) &   &  &  & SELF\_AGR& -4.38*** (0.000)   & -0.00 (0.954)   & 0.08 (0.414) \\
 & SELF\_CON& 1.72* (0.034)  & -0.03 (0.709)  & 0.06 (0.415) &   &  &  & SELF\_CON& -0.31 (0.569)  & 0.07 (0.132)& 0.08 (0.353) \\
 & SELF\_NEU& -1.10 (0.230)  & 0.07 (0.403)   & -0.18* (0.026) &   &  &  & SELF\_NEU& -1.02 (0.081)  & -0.02 (0.736)   & 0.00 (0.959) \\
 & SELF\_OPE& 1.94* (0.017)  & 0.08 (0.264)   & 0.00 (0.980) &   &  &  & SELF\_OPE& 0.08 (0.878)   & 0.07 (0.124)& -0.18* (0.049) \\
 & PARTNER\_EXT & 1.45 (0.147)   & -0.12 (0.200)  & 0.00 (0.974) &   &  &  & PARTNER\_EXT & -0.09 (0.873)  & -0.01 (0.806)   & -0.07 (0.499)  \\
 & PARTNER\_AGR & 0.58 (0.503)   & -0.04 (0.582)  & -0.14 (0.060)  &   &  &  & PARTNER\_AGR & 0.81 (0.147)   & 0.00 (0.954)& -0.08 (0.414)  \\
 & PARTNER\_CON & -0.05 (0.954)  & 0.03 (0.709)   & -0.06 (0.415)  &   &  &  & PARTNER\_CON & -0.90 (0.104)  & -0.07 (0.132)   & -0.08 (0.353)  \\
 & PARTNER\_NEU & -0.64 (0.486)  & -0.07 (0.403)  & 0.18* (0.026)&   &  &  & PARTNER\_NEU & -0.05 (0.936)  & 0.02 (0.736)& -0.00 (0.959)  \\
 & PARTNER\_OPE & -0.40 (0.615)  & -0.08 (0.264)  & -0.00 (0.980)  &   &  &  & PARTNER\_OPE & 0.27 (0.608)   & -0.07 (0.124)   & 0.18* (0.049)\\
 & POSITION & -5.31*** (0.000)   & -0.20 (0.134)  & -0.19 (0.151)  &   &  &  & POSITION & 16.85*** (0.000)   & -0.22* (0.010)  & 0.94*** (0.000)  \\ \cmidrule(r){1-5} \cmidrule(l){8-12} 
\multirow{12}{*}{Claude Sonnet 3.7}  & CONST& 55.29*** (0.000)   & 0.00 (1.000)   & N.A   &   &  & \multirow{12}{*}{KODIS}  & CONST& 60.20*** (0.000)   & -0.01 (0.927)   & -0.00 (1.000) \\
 & SELF\_EXT& 0.33 (0.607)   & -0.17** (0.005)& N.A   &   &  &  & SELF\_EXT& -2.18 (0.142)  & 0.05 (0.697)& 0.59 (0.147) \\
 & SELF\_AGR& -2.50*** (0.000)   & -0.08 (0.179)  & N.A   &   &  &  & SELF\_AGR& 0.69 (0.679)   & -0.10 (0.471)   & 0.53 (0.248) \\
 & SELF\_CON& 0.28 (0.629)   & 0.00 (0.962)   & N.A   &   &  &  & SELF\_CON& 2.02 (0.242)   & -0.25 (0.078)   & 0.15 (0.684) \\
 & SELF\_NEU& -0.01 (0.990)  & 0.03 (0.635)   & N.A   &   &  &  & SELF\_NEU& -0.14 (0.924)  & -0.26* (0.026)  & 0.52 (0.112) \\
 & SELF\_OPE& 0.64 (0.240)   & 0.00 (0.925)   & N.A   &   &  &  & SELF\_OPE& -1.04 (0.537)  & -0.01 (0.969)   & -0.15 (0.696)  \\
 & PARTNER\_EXT & -1.42* (0.028) & 0.17** (0.005) & N.A   &   &  &  & PARTNER\_EXT & 1.52 (0.305)   & -0.06 (0.616)   & -0.59 (0.147)  \\
 & PARTNER\_AGR & 3.05*** (0.000)& 0.08 (0.179)   & N.A   &   &  &  & PARTNER\_AGR & -0.01 (0.996)  & 0.09 (0.486)& -0.53 (0.248)  \\
 & PARTNER\_CON & 0.31 (0.602)   & -0.00 (0.962)  & N.A   &   &  &  & PARTNER\_CON & -1.93 (0.266)  & 0.26 (0.067)& -0.15 (0.684)  \\
 & PARTNER\_NEU & -0.31 (0.616)  & -0.03 (0.635)  & N.A   &   &  &  & PARTNER\_NEU & -1.85 (0.204)  & 0.27* (0.025)   & -0.52 (0.112)  \\
 & PARTNER\_OPE & 0.24 (0.657)   & -0.00 (0.925)  & N.A   &   &  &  & PARTNER\_OPE & -1.34 (0.426)  & 0.01 (0.914)& 0.15 (0.696) \\
 & POSITION & -12.05*** (0.000)  & -0.47*** (0.000)   & N.A   &   &  &  & POSITION & -3.21* (0.021) & 0.49*** (0.000) & -0.24 (0.435)  \\ \cmidrule(r){1-5} \cmidrule(l){8-12} 
\end{tabular}}
\caption{Summary of all regression results for personality predictors across LLMs (L2L) and human (KODIS) conversations, focusing on dependent variables related to Participation and Resolution Behaviors. Coefficients (B) are reported with p-values in parentheses. *, , *** indicate p $<$ .05, .01, and .001, respectively.}
\label{tab:regression_full_results_participation_resolution}
\end{table*}

\begin{table*}[]
\resizebox{\textwidth}{!}{%
\begin{tabular}{llllllllllllll}
\hline
\multicolumn{1}{c}{}  &  & \multicolumn{12}{c}{Dependent Variables - IRP Ratio}\\ \hline
\multicolumn{1}{c}{Dataset}  & \multicolumn{1}{c}{\begin{tabular}[c]{@{}c@{}}Independent \\ Variables\end{tabular}} & \multicolumn{1}{c}{\textit{Cooperative}} & \multicolumn{1}{c}{\textit{Competitive}} & \multicolumn{1}{c}{\textit{Neutral}} & \multicolumn{1}{c|}{\textit{Residual}}  & \multicolumn{1}{c}{\begin{tabular}[c]{@{}c@{}}Positive \\ Expectations\end{tabular}} & \multicolumn{1}{c}{\textit{Proposal}} & \multicolumn{1}{c}{\textit{Concession}} & \multicolumn{1}{c}{\textit{Interests}} & \multicolumn{1}{c}{\textit{Facts}} & \multicolumn{1}{c}{\textit{Proceural}} & \multicolumn{1}{c}{\textit{Power}} & \multicolumn{1}{c}{\textit{Rights}} \\ \hline
 & CONST   & \textit{80.76*** (0.000)}  & \textit{5.88 (0.271)}  & \textit{1.36 (0.327)}  & \multicolumn{1}{l|}{\textit{39.81*** (0.000)}} &N.A & \textit{39.55*** (0.000)}  & \textit{11.43* (0.017)}   & \textit{-4.23 (0.143)}   & \textit{1.10 (0.462)}& \textit{-0.02 (0.438)}   & \textit{6.06 (0.391)}& \textit{1.98 (0.465)} \\
 & SELF\_EXT  & \textit{0.10 (0.752)}  & \textit{1.07* (0.043)} & \textit{0.18 (0.251)}  & \multicolumn{1}{l|}{\textit{-1.04 (0.165)}}&N.A & \textit{2.14*** (0.001)}& \textit{0.68 (0.152)} & \textit{0.25 (0.352)}& \textit{0.09 (0.548)}& \textit{0.00 (0.375)}& \textit{1.02 (0.136)}& \textit{0.11 (0.604)} \\
 & SELF\_AGR  & \textit{0.32 (0.312)}  & \textit{-0.18 (0.641)} & \textit{0.07 (0.608)}  & \multicolumn{1}{l|}{\textit{-1.17 (0.071)}}&N.A & \textit{1.29* (0.012)}  & \textit{-0.44 (0.274)}& \textit{0.65** (0.005)}  & \textit{0.02 (0.848)}& \textit{-0.01 (0.323)}   & \textit{0.27 (0.584)}& \textit{0.25 (0.168)} \\
 & SELF\_CON  & \textit{0.39 (0.164)}  & \textit{-0.22 (0.613)} & \textit{0.03 (0.815)}  & \multicolumn{1}{l|}{\textit{-0.45 (0.454)}}&N.A & \textit{0.62 (0.202)}   & \textit{-0.29 (0.485)}& \textit{0.07 (0.776)}& \textit{0.13 (0.294)}& \textit{-0.00 (0.739)}   & \textit{-0.17 (0.752)}  & \textit{0.06 (0.765)} \\
 & SELF\_NEU  & \textit{-0.30 (0.299)} & \textit{0.23 (0.616)}  & \textit{0.02 (0.868)}  & \multicolumn{1}{l|}{\textit{-0.48 (0.456)}}&N.A & \textit{-0.30 (0.558)}  & \textit{0.39 (0.339)} & \textit{0.20 (0.453)}& \textit{0.02 (0.860)}& \textit{0.01 (0.321)}& \textit{-0.18 (0.749)}  & \textit{0.06 (0.748)} \\
 & SELF\_OPE  & \textit{-0.08 (0.762)} & \textit{0.19 (0.649)}  & \textit{-0.06 (0.638)} & \multicolumn{1}{l|}{\textit{-1.26* (0.039)}}   &N.A & \textit{0.22 (0.643)}   & \textit{-0.08 (0.834)}& \textit{0.28 (0.206)}& \textit{-0.07 (0.595)}  & \textit{-0.00 (0.828)}   & \textit{0.42 (0.380)}& \textit{-0.17 (0.439)}\\
 & PARTNER\_EXT   & \textit{0.22 (0.487)}  & \textit{1.09* (0.041)} & \textit{0.17 (0.293)}  & \multicolumn{1}{l|}{\textit{0.57 (0.466)}} &N.A & \textit{0.08 (0.877)}   & \textit{0.31 (0.487)} & \textit{0.33 (0.255)}& \textit{0.22 (0.141)}& \textit{0.01 (0.326)}& \textit{0.42 (0.489)}& \textit{-0.11 (0.607)}\\
 & PARTNER\_AGR   & \textit{0.23 (0.456)}  & \textit{-0.15 (0.693)} & \textit{0.07 (0.598)}  & \multicolumn{1}{l|}{\textit{-0.21 (0.735)}}&N.A & \textit{0.41 (0.412)}   & \textit{-0.16 (0.678)}& \textit{-0.18 (0.498)}   & \textit{0.00 (0.976)}& \textit{0.01 (0.325)}& \textit{0.50 (0.297)}& \textit{-0.07 (0.718)}\\
 & PARTNER\_CON   & \textit{0.34 (0.218)}  & \textit{-0.23 (0.603)} & \textit{0.05 (0.695)}  & \multicolumn{1}{l|}{\textit{0.14 (0.817)}} &N.A & \textit{-0.01 (0.990)}  & \textit{0.26 (0.494)} & \textit{0.45* (0.031)}   & \textit{-0.16 (0.228)}  & \textit{0.00 (0.712)}& \textit{0.10 (0.843)}& \textit{-0.05 (0.829)}\\
 & PARTNER\_NEU   & \textit{-0.15 (0.597)} & \textit{0.24 (0.604)}  & \textit{0.02 (0.882)}  & \multicolumn{1}{l|}{\textit{1.02 (0.121)}} &N.A & \textit{-0.35 (0.495)}  & \textit{-0.33 (0.404)}& \textit{0.24 (0.323)}& \textit{0.04 (0.762)}& \textit{-0.02 (0.320)}   & \textit{-0.45 (0.426)}  & \textit{-0.01 (0.963)}\\
 & PARTNER\_OPE   & \textit{0.03 (0.914)}  & \textit{0.14 (0.727)}  & \textit{-0.06 (0.630)} & \multicolumn{1}{l|}{\textit{-0.91 (0.165)}}&N.A & \textit{-0.30 (0.494)}  & \textit{0.27 (0.430)} & \textit{0.44* (0.040)}   & \textit{0.13 (0.248)}& \textit{0.01 (0.325)}& \textit{-0.07 (0.890)}  & \textit{0.15 (0.393)} \\
\multirow{-12}{*}{Gemini 2.0 Flash} & POSITION& \textit{1.88*** (0.000)}   & \textit{0.20 (0.784)}  & \textit{0.06 (0.806)}  & \multicolumn{1}{l|}{\textit{11.44*** (0.000)}} &N.A & \textit{-12.42*** (0.000)} & \textit{-3.75*** (0.000)} & \textit{-3.26*** (0.000)}& \textit{-0.43 (0.059)}  & \textit{-0.02 (0.321)}   & \textit{3.32*** (0.000)}& \textit{0.45 (0.218)} \\
\rowcolor[HTML]{DEDCDC} 
\cellcolor[HTML]{DEDCDC}  & CONST   & 81.01*** (0.000)& 9.59** (0.003)  & 17.43*** (0.000)& \multicolumn{1}{l|}{\cellcolor[HTML]{DEDCDC}8.40 (0.062)}& -1.68 (0.058)  & 34.84*** (0.000) & 27.79*** (0.000)   & 23.04*** (0.001)  & 18.31*** (0.000) & 0.30 (0.449)  & 11.73*** (0.001) & 8.10* (0.031)  \\
\rowcolor[HTML]{DEDCDC} 
\cellcolor[HTML]{DEDCDC}  & SELF\_EXT  & 0.20 (0.608) & -0.18 (0.486)& 0.56 (0.163)& \multicolumn{1}{l|}{\cellcolor[HTML]{DEDCDC}-0.29 (0.450)}   & 0.04 (0.566)   & 1.86*** (0.000)  & -0.16 (0.704)  & -0.90 (0.118) & 1.23** (0.001)& -0.04 (0.616) & 0.48 (0.091)  & 0.39 (0.217)   \\
\rowcolor[HTML]{DEDCDC} 
\cellcolor[HTML]{DEDCDC}  & SELF\_AGR  & 0.89* (0.013)& -0.70* (0.012)  & -0.28 (0.480)   & \multicolumn{1}{l|}{\cellcolor[HTML]{DEDCDC}0.02 (0.956)}& 0.07 (0.473)   & 1.83*** (0.000)  & -1.37*** (0.001)   & 2.99*** (0.000)   & -0.15 (0.696) & 0.03 (0.365)  & -1.75*** (0.000) & -0.75* (0.012) \\
\rowcolor[HTML]{DEDCDC} 
\cellcolor[HTML]{DEDCDC}  & SELF\_CON  & -0.30 (0.381)& -0.03 (0.905)& 0.30 (0.455)& \multicolumn{1}{l|}{\cellcolor[HTML]{DEDCDC}-0.56 (0.150)}   & 0.15* (0.022)  & 0.07 (0.887) & -0.35 (0.394)  & -0.55 (0.315) & 0.53 (0.193)  & -0.07 (0.257) & -0.06 (0.817) & -0.25 (0.402)  \\
\rowcolor[HTML]{DEDCDC} 
\cellcolor[HTML]{DEDCDC}  & SELF\_NEU  & -0.51 (0.166)& 0.37 (0.091) & -0.32 (0.450)   & \multicolumn{1}{l|}{\cellcolor[HTML]{DEDCDC}0.53 (0.173)}& 0.09 (0.334)   & 0.35 (0.495) & -0.38 (0.368)  & -0.49 (0.386) & -0.29 (0.467) & -0.06 (0.096) & 0.93*** (0.001)  & 0.11 (0.703)   \\
\rowcolor[HTML]{DEDCDC} 
\cellcolor[HTML]{DEDCDC}  & SELF\_OPE  & 0.27 (0.418) & 0.07 (0.761) & 0.71* (0.039)   & \multicolumn{1}{l|}{\cellcolor[HTML]{DEDCDC}-0.06 (0.868)}   & 0.09 (0.219)   & 0.25 (0.581) & -0.38 (0.313)  & 0.34 (0.495)  & 0.36 (0.291)  & -0.05 (0.242) & -0.12 (0.566) & -0.12 (0.619)  \\
\rowcolor[HTML]{DEDCDC} 
\cellcolor[HTML]{DEDCDC}  & PARTNER\_EXT   & -0.14 (0.724)& -0.21 (0.422)& 0.47 (0.228)& \multicolumn{1}{l|}{\cellcolor[HTML]{DEDCDC}-0.13 (0.748)}   & -0.04 (0.637)  & -0.89 (0.085)& 0.14 (0.754)& 0.59 (0.299)  & 0.01 (0.985)  & 0.03 (0.505)  & -0.07 (0.804) & -0.24 (0.439)  \\
\rowcolor[HTML]{DEDCDC} 
\cellcolor[HTML]{DEDCDC}  & PARTNER\_AGR   & 0.71 (0.051) & -0.79** (0.005) & -0.33 (0.393)   & \multicolumn{1}{l|}{\cellcolor[HTML]{DEDCDC}0.31 (0.383)}& -0.00 (0.954)  & -0.96* (0.045)   & 0.73 (0.069)& -1.17* (0.034)& 0.22 (0.551)  & 0.02 (0.545)  & -0.24 (0.272) & 0.12 (0.662)   \\
\rowcolor[HTML]{DEDCDC} 
\cellcolor[HTML]{DEDCDC}  & PARTNER\_CON   & -0.25 (0.471)& -0.02 (0.925)& 0.27 (0.504)& \multicolumn{1}{l|}{\cellcolor[HTML]{DEDCDC}0.76* (0.027)}   & 0.06 (0.410)   & 0.40 (0.395) & 0.18 (0.641)& -0.51 (0.331) & 0.18 (0.599)  & 0.02 (0.586)  & -0.50 (0.062) & 0.16 (0.566)   \\
\rowcolor[HTML]{DEDCDC} 
\cellcolor[HTML]{DEDCDC}  & PARTNER\_NEU   & -0.60 (0.097)& 0.35 (0.108) & -0.37 (0.368)   & \multicolumn{1}{l|}{\cellcolor[HTML]{DEDCDC}-0.16 (0.685)}   & 0.09 (0.319)   & -0.59 (0.258)& 0.65 (0.131)& 0.22 (0.697)  & -0.16 (0.703) & -0.02 (0.608) & 0.20 (0.404)  & 0.15 (0.593)   \\
\rowcolor[HTML]{DEDCDC} 
\cellcolor[HTML]{DEDCDC}  & PARTNER\_OPE   & 0.27 (0.414) & 0.05 (0.826) & 0.76* (0.024)   & \multicolumn{1}{l|}{\cellcolor[HTML]{DEDCDC}-0.06 (0.849)}   & 0.07 (0.334)   & 0.52 (0.255) & 0.15 (0.694)& 0.41 (0.392)  & 0.37 (0.254)  & 0.07* (0.016) & -0.07 (0.770) & -0.02 (0.945)  \\
\rowcolor[HTML]{DEDCDC} 
\multirow{-12}{*}{\cellcolor[HTML]{DEDCDC}Claude Sonnet 3.7} & POSITION& 2.96*** (0.000) & 0.14 (0.722) & 0.85 (0.196)& \multicolumn{1}{l|}{\cellcolor[HTML]{DEDCDC}3.65*** (0.000)} & 0.49*** (0.001)& 1.38 (0.094) & -0.67 (0.329)  & -3.96*** (0.000)  & 3.98*** (0.000)  & -0.09 (0.196) & 3.43*** (0.000)  & 2.72*** (0.000)\\
 & CONST   & 82.41*** (0.000)& 11.77** (0.008) & 13.64*** (0.000)& \multicolumn{1}{l|}{17.56*** (0.000)}   & 10.83*** (0.000)  & 42.68*** (0.000) & 23.37*** (0.000)   & 13.42*** (0.000)  & 1.78 (0.420)  & 5.09* (0.046) & 12.32*** (0.000) & 15.22** (0.002)\\
 & SELF\_EXT  & 0.41 (0.072) & 0.95* (0.025)& 0.82* (0.011)   & \multicolumn{1}{l|}{1.14** (0.003)} & 1.08*** (0.000)& 2.36*** (0.000)  & -0.06 (0.883)  & 0.35 (0.297)  & 0.42* (0.044) & 0.89*** (0.001)   & 0.80* (0.018) & 0.49 (0.294)   \\
 & SELF\_AGR  & 0.32 (0.120) & -0.22 (0.574)& 0.20 (0.543)& \multicolumn{1}{l|}{1.84*** (0.000)}& 0.84*** (0.000)& 1.18** (0.005)   & 0.79* (0.041)  & 2.41*** (0.000)   & 0.29 (0.149)  & 0.55* (0.022) & -1.36*** (0.000) & 0.12 (0.772)   \\
 & SELF\_CON  & 0.08 (0.702) & 0.59 (0.122) & 0.26 (0.406)& \multicolumn{1}{l|}{-0.13 (0.709)}  & -0.17 (0.431)  & -0.04 (0.915)& 0.79* (0.028)  & 0.17 (0.604)  & -0.04 (0.858) & 0.21 (0.343)  & -0.19 (0.509) & 0.62 (0.156)   \\
 & SELF\_NEU  & -0.04 (0.866)& 0.88* (0.030)& -0.23 (0.473)   & \multicolumn{1}{l|}{-1.40*** (0.000)}   & -0.62** (0.008)& -0.39 (0.363)& -0.21 (0.589)  & -0.76* (0.023)& -0.17 (0.425) & -0.29 (0.243) & 0.65* (0.034) & 0.31 (0.486)   \\
 & SELF\_OPE  & 0.24 (0.234) & -0.18 (0.649)& -0.12 (0.692)   & \multicolumn{1}{l|}{0.31 (0.377)}& 0.17 (0.431)   & 1.39*** (0.001)  & -0.25 (0.491)  & 0.72* (0.016) & 0.11 (0.547)  & 0.16 (0.499)  & -0.55 (0.063) & 0.31 (0.482)   \\
 & PARTNER\_EXT   & 0.50* (0.031)& 1.02* (0.017)& 0.89** (0.006)  & \multicolumn{1}{l|}{-0.30 (0.478)}  & -0.66** (0.008)& -0.69 (0.126)& 0.62 (0.125)& 0.19 (0.567)  & 0.07 (0.757)  & -0.09 (0.724) & 0.12 (0.716)  & 1.43** (0.002) \\
 & PARTNER\_AGR   & 0.21 (0.325) & -0.27 (0.495)& 0.06 (0.846)& \multicolumn{1}{l|}{0.30 (0.391)}& -0.06 (0.788)  & 0.14 (0.734) & 0.77* (0.040)  & -0.39 (0.220) & 0.30 (0.145)  & -0.13 (0.608) & -0.10 (0.702) & 0.12 (0.778)   \\
 & PARTNER\_CON   & 0.09 (0.683) & 0.62 (0.108) & 0.26 (0.401)& \multicolumn{1}{l|}{0.16 (0.650)}& -0.27 (0.215)  & -0.34 (0.396)& 0.40 (0.270)& -0.10 (0.729) & 0.24 (0.206)  & 0.43 (0.066)  & 0.29 (0.324)  & -0.26 (0.532)  \\
 & PARTNER\_NEU   & -0.02 (0.946)& 0.90* (0.028)& -0.17 (0.592)   & \multicolumn{1}{l|}{-0.10 (0.801)}  & -0.13 (0.595)  & -0.39 (0.369)& 0.16 (0.687)& -0.12 (0.721) & 0.24 (0.226)  & -0.03 (0.918) & -0.14 (0.662) & 0.75 (0.098)   \\
 & PARTNER\_OPE   & 0.20 (0.314) & -0.15 (0.706)& -0.16 (0.599)   & \multicolumn{1}{l|}{0.11 (0.745)}& -0.44 (0.060)  & -0.04 (0.916)& 0.49 (0.177)& 0.25 (0.395)  & -0.15 (0.464) & -0.25 (0.291) & -0.22 (0.429) & 0.03 (0.952)   \\
\multirow{-12}{*}{GPT-4o mini}   & POSITION& 1.57*** (0.000) & 0.50 (0.481) & 0.31 (0.575)& \multicolumn{1}{l|}{-1.36* (0.035)} & -3.14*** (0.000)  & -15.91*** (0.000)& 5.34*** (0.000)& -8.67*** (0.000)  & -0.57 (0.109) & -3.09*** (0.000)  & 4.32*** (0.000)  & 7.58*** (0.000)\\
\rowcolor[HTML]{DEDCDC} 
\cellcolor[HTML]{DEDCDC}  & CONST   & 60.29*** (0.000)& 11.06 (0.255)& 50.88*** (0.000)& \multicolumn{1}{l|}{\cellcolor[HTML]{DEDCDC}16.08 (0.144)}   & 13.74* (0.024) & 45.95*** (0.000) & -0.49 (0.960)  & 12.85 (0.243) & 25.97* (0.038)& 26.80*** (0.000)  & 9.11 (0.224)  & 12.06 (0.249)  \\
\rowcolor[HTML]{DEDCDC} 
\cellcolor[HTML]{DEDCDC}  & SELF\_EXT  & 0.70 (0.529) & 1.74* (0.024)& 0.20 (0.854)& \multicolumn{1}{l|}{\cellcolor[HTML]{DEDCDC}0.65 (0.515)}& -0.23 (0.662)  & 0.81 (0.448) & 0.04 (0.962)& 0.88 (0.321)  & 0.64 (0.531)  & -0.39 (0.495) & 1.02 (0.129)  & 0.01 (0.994)   \\
\rowcolor[HTML]{DEDCDC} 
\cellcolor[HTML]{DEDCDC}  & SELF\_AGR  & 0.76 (0.542) & 0.76 (0.368) & -1.52 (0.194)   & \multicolumn{1}{l|}{\cellcolor[HTML]{DEDCDC}-1.82 (0.106)}   & 0.25 (0.704)   & -1.46 (0.183)& -0.20 (0.827)  & 0.73 (0.455)  & 0.34 (0.758)  & -0.91 (0.247) & -0.11 (0.869) & -0.41 (0.710)  \\
\rowcolor[HTML]{DEDCDC} 
\cellcolor[HTML]{DEDCDC}  & SELF\_CON  & -1.52 (0.275)& -1.61 (0.069)& 0.61 (0.656)& \multicolumn{1}{l|}{\cellcolor[HTML]{DEDCDC}1.40 (0.195)}& -0.92 (0.156)  & -0.39 (0.766)& 0.17 (0.851)& -0.63 (0.521) & 1.07 (0.385)  & -0.32 (0.691) & -0.99 (0.214) & -0.27 (0.807)  \\
\rowcolor[HTML]{DEDCDC} 
\cellcolor[HTML]{DEDCDC}  & SELF\_NEU  & -0.33 (0.759)& 1.11 (0.124) & 0.53 (0.602)& \multicolumn{1}{l|}{\cellcolor[HTML]{DEDCDC}-0.72 (0.447)}   & -1.33* (0.019) & 0.30 (0.758) & 0.67 (0.369)& 0.04 (0.960)  & 0.96 (0.320)  & -1.07 (0.082) & 0.50 (0.449)  & 0.16 (0.852)   \\
\rowcolor[HTML]{DEDCDC} 
\cellcolor[HTML]{DEDCDC}  & SELF\_OPE  & 1.02 (0.396) & -0.84 (0.403)& -1.09 (0.362)   & \multicolumn{1}{l|}{\cellcolor[HTML]{DEDCDC}-0.86 (0.442)}   & 1.43* (0.021)  & 1.69 (0.138) & -0.48 (0.616)  & 0.51 (0.607)  & -1.03 (0.380) & 0.82 (0.250)  & 0.17 (0.820)  & 0.46 (0.683)   \\
\rowcolor[HTML]{DEDCDC} 
\cellcolor[HTML]{DEDCDC}  & PARTNER\_EXT   & 0.56 (0.605) & 1.80* (0.019)& 0.28 (0.794)& \multicolumn{1}{l|}{\cellcolor[HTML]{DEDCDC}-0.91 (0.366)}   & -0.60 (0.215)  & -0.08 (0.940)& 1.48 (0.068)& 0.35 (0.708)  & 1.49 (0.115)  & -0.90 (0.136) & 1.13 (0.084)  & 0.44 (0.626)   \\
\rowcolor[HTML]{DEDCDC} 
\cellcolor[HTML]{DEDCDC}  & PARTNER\_AGR   & 0.45 (0.724) & 0.52 (0.543) & -1.60 (0.157)   & \multicolumn{1}{l|}{\cellcolor[HTML]{DEDCDC}0.36 (0.747)}& 0.06 (0.921)   & 0.48 (0.657) & 1.45 (0.105)& -0.09 (0.936) & -1.38 (0.211) & -0.35 (0.643) & -0.07 (0.912) & 0.62 (0.519)   \\
\rowcolor[HTML]{DEDCDC} 
\cellcolor[HTML]{DEDCDC}  & PARTNER\_CON   & -1.84 (0.175)& -1.78* (0.046)  & 0.29 (0.831)& \multicolumn{1}{l|}{\cellcolor[HTML]{DEDCDC}2.32* (0.030)}   & -0.48 (0.401)  & -1.97 (0.093)& -1.16 (0.221)  & 0.03 (0.974)  & 1.21 (0.330)  & 0.26 (0.720)  & -0.20 (0.796) & -1.83* (0.049) \\
\rowcolor[HTML]{DEDCDC} 
\cellcolor[HTML]{DEDCDC}  & PARTNER\_NEU   & -0.71 (0.512)& 1.02 (0.153) & 0.13 (0.896)& \multicolumn{1}{l|}{\cellcolor[HTML]{DEDCDC}1.16 (0.191)}& -0.67 (0.197)  & -2.32* (0.022)   & 1.82* (0.016)  & -0.77 (0.389) & 1.02 (0.292)  & -1.36* (0.013)& 0.16 (0.784)  & 0.44 (0.571)   \\
\rowcolor[HTML]{DEDCDC} 
\cellcolor[HTML]{DEDCDC}  & PARTNER\_OPE   & 0.77 (0.518) & -0.84 (0.399)& -1.24 (0.287)   & \multicolumn{1}{l|}{\cellcolor[HTML]{DEDCDC}0.63 (0.551)}& -0.35 (0.578)  & -0.94 (0.421)& 1.75* (0.044)  & -0.04 (0.968) & -1.96 (0.089) & -1.47* (0.033)& -1.32 (0.119) & 0.92 (0.321)   \\
\rowcolor[HTML]{DEDCDC} 
\multirow{-12}{*}{\cellcolor[HTML]{DEDCDC}KODIS}  & POSITION& 1.88 (0.063) & 0.37 (0.621) & 1.17 (0.213)& \multicolumn{1}{l|}{\cellcolor[HTML]{DEDCDC}0.72 (0.436)}& 2.27*** (0.000)& -6.54*** (0.000) & 1.62* (0.033)  & -1.91* (0.023)& 1.04 (0.260)  & 0.90 (0.107)  & -1.22* (0.050)& 8.29*** (0.000)\\ \hline
\end{tabular}%
}
\caption{Summary of all regression results for personality predictors across LLM (L2L) and human (KODIS) conversations, presented by dependent variables for IRP Ratio. Coefficients (B) are reported with p-values in parentheses. *, **, *** indicate p \textless .05, .01, and .001, respectively. Results are from Ordinary Least Squares (OLS) regression with robust standard errors (SE)}
\label{tab:irp_full_regressions_ratio}
\end{table*}

\begin{table*}[]
\resizebox{\textwidth}{!}{%
\begin{tabular}{llllllllllllll}
\hline
\multicolumn{1}{c}{}  &   & \multicolumn{12}{c}{Dependent Variables - IRP Reciprocity}\\ \hline
\multicolumn{1}{c}{Dataset}  & \begin{tabular}[c]{@{}l@{}}Independent \\ Variables\end{tabular} & \textit{Cooperative}& \textit{Competitive} & \textit{Neutral} & \multicolumn{1}{l|}{\textit{Residual}}  & \begin{tabular}[c]{@{}l@{}}Positive \\ Expectations\end{tabular} & \textit{Proposal}  & \textit{Concession}  & \textit{Interests}  & \textit{Facts}   & \textit{Proceural} & \textit{Power} & \textit{Rights} \\ \hline
 & CONST& \textit{58.18*** (0.000)}  & \textit{1.93 (0.897)}& \textit{-6.43 (0.642)}  & \multicolumn{1}{l|}{\textit{43.73** (0.003)}}  &N.A  & \textit{28.71** (0.001)}  & \textit{11.10 (0.442)}   & \textit{14.88 (0.446)}  & \textit{-6.44 (0.642)}  &N.A  & \textit{-6.88 (0.652)}& \textit{8.42 (0.633)}  \\
 & SELF\_EXT& \textit{3.08*** (0.000)}   & \textit{1.98 (0.200)}& \textit{-0.90 (0.575)}  & \multicolumn{1}{l|}{\textit{-0.83 (0.498)}}&N.A  & \textit{2.77** (0.002)}   & \textit{0.32 (0.822)}& \textit{0.26 (0.879)}   & \textit{-0.90 (0.576)}  &N.A  & \textit{3.46* (0.043)}& \textit{0.16 (0.946)}  \\
 & SELF\_AGR& \textit{1.51** (0.002)}& \textit{1.31 (0.273)}& \textit{-0.99 (0.289)}  & \multicolumn{1}{l|}{\textit{-0.79 (0.475)}}&N.A  & \textit{1.99** (0.006)}   & \textit{-0.70 (0.588)}   & \textit{2.61 (0.074)}   & \textit{-0.99 (0.292)}  &N.A  & \textit{1.45 (0.285)} & \textit{0.18 (0.886)}  \\
 & SELF\_CON& \textit{0.15 (0.746)}  & \textit{0.93 (0.455)}& \textit{1.56 (0.120)}   & \multicolumn{1}{l|}{\textit{-0.04 (0.969)}}&N.A  & \textit{0.68 (0.359)} & \textit{0.99 (0.433)}& \textit{-0.77 (0.624)}  & \textit{1.56 (0.120)}   &N.A  & \textit{0.10 (0.946)} & \textit{1.43 (0.455)}  \\
 & SELF\_NEU& \textit{0.19 (0.705)}  & \textit{0.19 (0.879)}& \textit{-0.28 (0.855)}  & \multicolumn{1}{l|}{\textit{-0.52 (0.628)}}&N.A  & \textit{-0.64 (0.415)}& \textit{1.56 (0.247)}& \textit{0.39 (0.823)}   & \textit{-0.28 (0.855)}  &N.A  & \textit{-0.20 (0.882)}& \textit{0.82 (0.668)}  \\
 & SELF\_OPE& \textit{0.68 (0.132)}  & \textit{-0.77 (0.545)}   & \textit{1.07 (0.397)}   & \multicolumn{1}{l|}{\textit{-1.10 (0.281)}}&N.A  & \textit{0.52 (0.461)} & \textit{0.60 (0.576)}& \textit{0.45 (0.774)}   & \textit{1.08 (0.397)}   &N.A  & \textit{0.90 (0.505)} & \textit{-1.91 (0.358)} \\
 & PARTNER\_EXT& \textit{-0.15 (0.770)} & \textit{0.33 (0.815)}& \textit{2.76 (0.139)}   & \multicolumn{1}{l|}{\textit{1.61 (0.225)}} &N.A  & \textit{0.35 (0.666)} & \textit{-1.06 (0.442)}   & \textit{-0.82 (0.629)}  & \textit{2.76 (0.139)}   &N.A  & \textit{0.47 (0.768)} & \textit{-0.89 (0.591)} \\
 & PARTNER\_AGR& \textit{-0.41 (0.335)} & \textit{-0.19 (0.886)}   & \textit{1.63 (0.295)}   & \multicolumn{1}{l|}{\textit{0.77 (0.456)}} &N.A  & \textit{0.75 (0.321)} & \textit{2.40* (0.033)}   & \textit{-4.46* (0.026)} & \textit{1.63 (0.295)}   &N.A  & \textit{1.89 (0.176)} & \textit{-3.01 (0.213)} \\
 & PARTNER\_CON& \textit{0.28 (0.534)}  & \textit{1.35 (0.271)}& \textit{1.40 (0.446)}   & \multicolumn{1}{l|}{\textit{0.37 (0.731)}} &N.A  & \textit{0.67 (0.341)} & \textit{-0.44 (0.702)}   & \textit{0.17 (0.906)}   & \textit{1.40 (0.448)}   &N.A  & \textit{1.45 (0.276)} & \textit{2.15 (0.138)}  \\
 & PARTNER\_NEU& \textit{-0.62 (0.178)} & \textit{0.36 (0.802)}& \textit{-0.22 (0.855)}  & \multicolumn{1}{l|}{\textit{0.43 (0.701)}} &N.A  & \textit{-0.62 (0.413)}& \textit{-1.49 (0.230)}   & \textit{1.79 (0.226)}   & \textit{-0.22 (0.860)}  &N.A  & \textit{-0.31 (0.841)}& \textit{0.42 (0.819)}  \\
 & PARTNER\_OPE& \textit{-0.06 (0.899)} & \textit{1.44 (0.245)}& \textit{-2.02 (0.154)}  & \multicolumn{1}{l|}{\textit{-0.62 (0.549)}}&N.A  & \textit{0.28 (0.687)} & \textit{1.24 (0.284)}& \textit{2.17 (0.099)}   & \textit{-2.02 (0.154)}  &N.A  & \textit{0.29 (0.827)} & \textit{1.34 (0.458)}  \\
\multirow{-12}{*}{Gemini 2.0 Flash} & POSITION & \textit{-17.18*** (0.000)} & \textit{7.75*** (0.001)} & \textit{-5.34* (0.024)} & \multicolumn{1}{l|}{\textit{22.43*** (0.000)}} &N.A  & \textit{-9.84*** (0.000)} & \textit{-6.73** (0.003)} & \textit{-6.11 (0.067)}  & \textit{-5.34* (0.025)} &N.A  & \textit{12.33*** (0.000)} & \textit{-2.77 (0.339)} \\
\rowcolor[HTML]{DEDCDC} 
\cellcolor[HTML]{DEDCDC}  & CONST& 88.08*** (0.000)& 33.23* (0.030)& 41.50*** (0.000) & \multicolumn{1}{l|}{\cellcolor[HTML]{DEDCDC}28.16 (0.095)}   & 41.71 (0.516)   & 48.41*** (0.000)   & 64.74*** (0.000)  & 18.48 (0.182)& 42.90*** (0.000) & N.A & 37.02 (0.060)  & 44.18* (0.018)  \\
\rowcolor[HTML]{DEDCDC} 
\cellcolor[HTML]{DEDCDC}  & SELF\_EXT& 0.23 (0.457) & 1.62 (0.162)  & 2.82** (0.002)   & \multicolumn{1}{l|}{\cellcolor[HTML]{DEDCDC}-1.78 (0.178)}   & 2.67 (0.589)& 3.36*** (0.000)& -0.60 (0.669) & 0.33 (0.779) & 2.90** (0.001)   & N.A & 0.76 (0.611)& 0.73 (0.645)\\
\rowcolor[HTML]{DEDCDC} 
\cellcolor[HTML]{DEDCDC}  & SELF\_AGR& 1.04*** (0.001) & -3.39** (0.004)   & -0.84 (0.331)& \multicolumn{1}{l|}{\cellcolor[HTML]{DEDCDC}-1.59 (0.238)}   & -4.09 (0.628)   & 1.56 (0.057)& -0.82 (0.532) & 4.27*** (0.000)  & -0.79 (0.367)& N.A & -2.35 (0.091)  & -2.57 (0.105)   \\
\rowcolor[HTML]{DEDCDC} 
\cellcolor[HTML]{DEDCDC}  & SELF\_CON& -0.10 (0.699)& 1.36 (0.209)  & 2.56** (0.004)   & \multicolumn{1}{l|}{\cellcolor[HTML]{DEDCDC}-1.63 (0.248)}   & 7.39 (0.410)& -0.40 (0.614)  & -1.21 (0.339) & -0.22 (0.832)& 2.39** (0.007)   & N.A & 0.70 (0.613)& -0.37 (0.804)   \\
\rowcolor[HTML]{DEDCDC} 
\cellcolor[HTML]{DEDCDC}  & SELF\_NEU& -0.24 (0.444)& 2.96* (0.019) & 0.17 (0.849) & \multicolumn{1}{l|}{\cellcolor[HTML]{DEDCDC}-0.26 (0.817)}   & 2.90 (0.699)& 0.97 (0.251)& -1.54 (0.265) & -0.62 (0.560)& 0.18 (0.846) & N.A & 0.72 (0.666)& 0.15 (0.904)\\
\rowcolor[HTML]{DEDCDC} 
\cellcolor[HTML]{DEDCDC}  & SELF\_OPE& -0.16 (0.566)& -0.51 (0.566) & -0.61 (0.430)& \multicolumn{1}{l|}{\cellcolor[HTML]{DEDCDC}0.62 (0.563)}& -4.97 (0.498)   & 1.52* (0.045)  & -0.58 (0.616) & 0.27 (0.787) & -0.55 (0.483)& N.A & -0.24 (0.874)  & -2.09* (0.025)  \\
\rowcolor[HTML]{DEDCDC} 
\cellcolor[HTML]{DEDCDC}  & PARTNER\_EXT& -0.36 (0.291)& -1.43 (0.219) & 0.58 (0.538) & \multicolumn{1}{l|}{\cellcolor[HTML]{DEDCDC}-1.10 (0.422)}   & -5.35 (0.396)   & -1.81* (0.041) & -0.68 (0.624) & 2.11 (0.069) & 0.46 (0.631) & N.A & -0.33 (0.840)  & -0.09 (0.947)   \\
\rowcolor[HTML]{DEDCDC} 
\cellcolor[HTML]{DEDCDC}  & PARTNER\_AGR& -0.87** (0.001) & -1.67 (0.135) & 0.86 (0.326) & \multicolumn{1}{l|}{\cellcolor[HTML]{DEDCDC}1.06 (0.310)}& -0.45 (0.974)   & -2.10* (0.013) & 2.50* (0.042) & -0.70 (0.522)& 0.80 (0.358) & N.A & -1.04 (0.431)  & -0.75 (0.547)   \\
\rowcolor[HTML]{DEDCDC} 
\cellcolor[HTML]{DEDCDC}  & PARTNER\_CON& 0.00 (0.995) & -1.71 (0.156) & -0.59 (0.510)& \multicolumn{1}{l|}{\cellcolor[HTML]{DEDCDC}-1.62 (0.199)}   & -0.36 (0.972)   & 0.06 (0.944)& -1.24 (0.319) & -0.46 (0.653)& -0.61 (0.501)& N.A & -0.35 (0.776)  & -1.06 (0.421)   \\
\rowcolor[HTML]{DEDCDC} 
\cellcolor[HTML]{DEDCDC}  & PARTNER\_NEU& 0.26 (0.394) & -2.78* (0.037)& 1.14 (0.200) & \multicolumn{1}{l|}{\cellcolor[HTML]{DEDCDC}1.28 (0.292)}& -4.42 (0.609)   & -1.65 (0.055)  & 0.67 (0.625)  & -0.30 (0.794)& 1.04 (0.243) & N.A & -3.65* (0.018) & -3.81** (0.005) \\
\rowcolor[HTML]{DEDCDC} 
\cellcolor[HTML]{DEDCDC}  & PARTNER\_OPE& 0.59* (0.036)& 1.71 (0.089)  & 0.41 (0.614) & \multicolumn{1}{l|}{\cellcolor[HTML]{DEDCDC}1.97 (0.080)}& -1.70 (0.867)   & 1.11 (0.141)& 0.52 (0.659)  & 0.33 (0.738) & 0.32 (0.697) & N.A & -0.38 (0.762)  & 1.48 (0.174)\\
\rowcolor[HTML]{DEDCDC} 
\multirow{-12}{*}{\cellcolor[HTML]{DEDCDC}Claude Sonnet 3.7} & POSITION & -5.99*** (0.000)& 9.44*** (0.000)   & 10.44*** (0.000) & \multicolumn{1}{l|}{\cellcolor[HTML]{DEDCDC}8.41*** (0.001)} & 9.79 (0.635)& 0.88 (0.523)& -4.00 (0.065) & -3.24 (0.078)& 10.39*** (0.000) & N.A & 8.79* (0.040)  & 6.66* (0.022)   \\
 & CONST& 85.90*** (0.000)& 41.39*** (0.000)  & 21.22* (0.026)   & \multicolumn{1}{l|}{33.16*** (0.000)}   & 20.47* (0.045)  & 44.14*** (0.000)   & 38.86*** (0.000)  & 19.60** (0.010)  & 25.31* (0.041)   & 10.89 (0.341)  & 35.52*** (0.001)   & 18.29* (0.028)  \\
 & SELF\_EXT& 0.37 (0.262) & 1.40* (0.049) & 1.26 (0.156) & \multicolumn{1}{l|}{1.11 (0.166)}& 1.86* (0.042)   & 1.80** (0.002) & -0.15 (0.823) & -0.48 (0.504)& 0.09 (0.936) & 2.19* (0.031)  & -0.10 (0.914)  & 0.93 (0.234)\\
 & SELF\_AGR& 1.10*** (0.000) & -1.47* (0.034)& 1.57* (0.048)& \multicolumn{1}{l|}{1.93** (0.008)} & 1.10 (0.200)& 1.28* (0.024)  & 0.68 (0.300)  & 2.81*** (0.000)  & -0.14 (0.891)& 0.92 (0.292)& -2.87** (0.003)& 1.23 (0.098)\\
 & SELF\_CON& 0.40 (0.171) & 0.38 (0.571)  & -0.14 (0.861)& \multicolumn{1}{l|}{-1.12 (0.125)}  & -0.36 (0.659)   & 0.20 (0.708)& 0.77 (0.228)  & 1.96** (0.002)   & -0.09 (0.928)& 1.30 (0.158)& -0.45 (0.596)  & 0.90 (0.223)\\
 & SELF\_NEU& -0.49 (0.123)& 1.41* (0.042) & 0.55 (0.516) & \multicolumn{1}{l|}{-0.39 (0.604)}  & 0.48 (0.586)& -0.18 (0.756)  & -0.83 (0.230) & -1.14 (0.087)& 0.27 (0.798) & 0.83 (0.387)& 0.15 (0.868)& 0.82 (0.278)\\
 & SELF\_OPE& 0.27 (0.374) & 0.09 (0.894)  & 0.67 (0.397) & \multicolumn{1}{l|}{-0.28 (0.685)}  & 1.33 (0.091)& 1.33* (0.011)  & -0.63 (0.327) & 0.45 (0.474) & 1.13 (0.257) & 1.31 (0.138)& -0.72 (0.374)  & -0.17 (0.813)   \\
 & PARTNER\_EXT& -0.89* (0.013)  & 0.75 (0.286)  & -2.08* (0.012)   & \multicolumn{1}{l|}{-0.73 (0.353)}  & -2.23* (0.018)  & -0.33 (0.584)  & 0.34 (0.644)  & 0.56 (0.425) & -1.44 (0.153)& -0.65 (0.515)  & -0.30 (0.745)  & 1.32 (0.093)\\
 & PARTNER\_AGR& 0.28 (0.315) & 0.61 (0.366)  & 0.06 (0.938) & \multicolumn{1}{l|}{0.51 (0.480)}& 0.07 (0.942)& 0.49 (0.384)& 1.80** (0.006)& -0.50 (0.463)& -0.72 (0.495)& -1.31 (0.187)  & 0.10 (0.906)& 0.96 (0.203)\\
 & PARTNER\_CON& -0.16 (0.593)& -1.42* (0.038)& 0.17 (0.836) & \multicolumn{1}{l|}{-0.79 (0.261)}  & -1.25 (0.131)   & -0.77 (0.159)  & 0.40 (0.542)  & -0.82 (0.196)& -0.71 (0.484)& -0.56 (0.561)  & -0.94 (0.300)  & -1.15 (0.117)   \\
 & PARTNER\_NEU& -0.18 (0.564)& -0.53 (0.454) & 1.65 (0.056) & \multicolumn{1}{l|}{-0.27 (0.730)}  & -1.35 (0.138)   & -0.24 (0.676)  & 0.22 (0.747)  & 0.56 (0.413) & 1.45 (0.185) & 1.71 (0.094)& -1.74 (0.074)  & -0.44 (0.562)   \\
 & PARTNER\_OPE& -0.03 (0.912)& 0.13 (0.843)  & -1.37 (0.073)& \multicolumn{1}{l|}{-0.37 (0.589)}  & -0.25 (0.743)   & 0.34 (0.518)& 1.15 (0.070)  & -0.07 (0.910)& -0.83 (0.394)& -1.48 (0.099)  & 1.84* (0.035)  & 0.22 (0.760)\\
\multirow{-12}{*}{GPT-4o mini}   & POSITION & -6.59*** (0.000)& 13.28*** (0.000)  & -3.78* (0.012)   & \multicolumn{1}{l|}{-2.51 (0.051)}  & -2.24 (0.151)   & -16.54*** (0.000)  & 3.51** (0.002)& -13.75*** (0.000)& -2.37 (0.194)& -9.71*** (0.000)   & 8.95*** (0.000)& 8.70*** (0.000) \\
\rowcolor[HTML]{DEDCDC} 
\cellcolor[HTML]{DEDCDC}  & CONST& 57.31** (0.004) & 37.10 (0.277) & 64.62** (0.009)  & \multicolumn{1}{l|}{\cellcolor[HTML]{DEDCDC}11.75 (0.730)}   & 42.83 (0.252)   & 67.95** (0.007)& 25.03 (0.506) & -21.15 (0.524)   & 50.88 (0.063)& 59.33 (0.084)  & 43.31 (0.412)  & 19.94 (0.510)   \\
\rowcolor[HTML]{DEDCDC} 
\cellcolor[HTML]{DEDCDC}  & SELF\_EXT& 0.10 (0.949) & 0.20 (0.945)  & 0.65 (0.751) & \multicolumn{1}{l|}{\cellcolor[HTML]{DEDCDC}3.00 (0.304)}& -4.78 (0.117)   & -0.05 (0.983)  & -2.71 (0.385) & 0.52 (0.849) & 0.74 (0.741) & -0.78 (0.816)  & 0.07 (0.987)& -1.87 (0.527)   \\
\rowcolor[HTML]{DEDCDC} 
\cellcolor[HTML]{DEDCDC}  & SELF\_AGR& -0.74 (0.701)& -3.12 (0.325) & 0.73 (0.742) & \multicolumn{1}{l|}{\cellcolor[HTML]{DEDCDC}-1.84 (0.559)}   & -4.73 (0.214)   & -0.06 (0.980)  & -0.08 (0.982) & 5.48 (0.054) & 1.10 (0.669) & 1.64 (0.621)& -5.90 (0.186)  & 0.70 (0.773)\\
\rowcolor[HTML]{DEDCDC} 
\cellcolor[HTML]{DEDCDC}  & SELF\_CON& 0.01 (0.995) & -0.68 (0.832) & -1.45 (0.535)& \multicolumn{1}{l|}{\cellcolor[HTML]{DEDCDC}0.02 (0.996)}& -2.89 (0.333)   & 1.06 (0.681)& 3.73 (0.298)  & 2.38 (0.421) & -0.19 (0.944)& -3.48 (0.382)  & 2.97 (0.532)& -0.24 (0.937)   \\
\rowcolor[HTML]{DEDCDC} 
\cellcolor[HTML]{DEDCDC}  & SELF\_NEU& 1.06 (0.521) & 2.45 (0.356)  & 0.41 (0.820) & \multicolumn{1}{l|}{\cellcolor[HTML]{DEDCDC}-2.45 (0.346)}   & -2.76 (0.208)   & 0.08 (0.968)& 5.21 (0.084)  & 4.14 (0.093) & 1.39 (0.475) & 0.25 (0.930)& 1.53 (0.681)& 1.48 (0.522)\\
\rowcolor[HTML]{DEDCDC} 
\cellcolor[HTML]{DEDCDC}  & SELF\_OPE& 4.03* (0.039)& 0.31 (0.927)  & 1.72 (0.463) & \multicolumn{1}{l|}{\cellcolor[HTML]{DEDCDC}-1.76 (0.600)}   & 1.12 (0.578)& 3.71 (0.120)& -6.58* (0.042)& 2.87 (0.344) & 1.42 (0.577) & 6.12 (0.141)& -2.31 (0.647)  & 2.80 (0.296)\\
\rowcolor[HTML]{DEDCDC} 
\cellcolor[HTML]{DEDCDC}  & PARTNER\_EXT& 2.24 (0.200) & 0.87 (0.764)  & 0.76 (0.717) & \multicolumn{1}{l|}{\cellcolor[HTML]{DEDCDC}0.56 (0.844)}& -4.28 (0.080)   & -0.32 (0.880)  & 5.58 (0.064)  & -1.51 (0.601)& 0.80 (0.724) & -6.34 (0.058)  & -2.86 (0.463)  & 1.57 (0.506)\\
\rowcolor[HTML]{DEDCDC} 
\cellcolor[HTML]{DEDCDC}  & PARTNER\_AGR& 2.41 (0.195) & 4.86 (0.157)  & -2.96 (0.176)& \multicolumn{1}{l|}{\cellcolor[HTML]{DEDCDC}0.87 (0.798)}& -0.66 (0.847)   & -2.89 (0.238)  & 1.37 (0.685)  & 2.92 (0.425) & -3.25 (0.201)& 1.11 (0.773)& 3.29 (0.516)& 3.52 (0.182)\\
\rowcolor[HTML]{DEDCDC} 
\cellcolor[HTML]{DEDCDC}  & PARTNER\_CON& -4.98** (0.008) & -4.32 (0.171) & 1.14 (0.641) & \multicolumn{1}{l|}{\cellcolor[HTML]{DEDCDC}4.66 (0.164)}& 3.53 (0.279)& -3.44 (0.180)  & -5.17 (0.139) & -2.82 (0.427)& 2.06 (0.449) & 4.48 (0.191)& 1.77 (0.675)& -9.32** (0.002) \\
\rowcolor[HTML]{DEDCDC} 
\cellcolor[HTML]{DEDCDC}  & PARTNER\_NEU& 0.16 (0.922) & -1.01 (0.714) & -0.13 (0.944)& \multicolumn{1}{l|}{\cellcolor[HTML]{DEDCDC}-0.57 (0.828)}   & -1.10 (0.632)   & -5.09* (0.013) & 0.48 (0.875)  & 1.07 (0.695) & 0.69 (0.733) & -4.20 (0.181)  & -0.46 (0.895)  & -2.60 (0.342)   \\
\rowcolor[HTML]{DEDCDC} 
\cellcolor[HTML]{DEDCDC}  & PARTNER\_OPE& 0.09 (0.959) & 0.81 (0.790)  & -1.64 (0.489)& \multicolumn{1}{l|}{\cellcolor[HTML]{DEDCDC}6.13* (0.048)}   & 3.11 (0.242)& -2.88 (0.250)  & 2.82 (0.390)  & -1.89 (0.548)& -2.22 (0.390)& -8.58** (0.010)& -4.84 (0.263)  & 3.65 (0.174)\\
\rowcolor[HTML]{DEDCDC} 
\multirow{-12}{*}{\cellcolor[HTML]{DEDCDC}KODIS}  & POSITION & -3.80* (0.015)  & 8.36** (0.004)& 3.94* (0.037)& \multicolumn{1}{l|}{\cellcolor[HTML]{DEDCDC}1.53 (0.550)}& 1.94 (0.487)& -5.41** (0.009)& 0.50 (0.861)  & 1.40 (0.584) & 1.58 (0.446) & 24.63*** (0.000)   & -2.75 (0.447)  & 8.78* (0.010)   \\ \cline{2-14} 
\end{tabular}%
}
\caption{Summary of all regression results for personality predictors across LLM (L2L) and human (KODIS) conversations, presented by dependent variables for IRP Reciprocity. Coefficients (B) are reported with p-values in parentheses. *, **, *** indicate p \textless .05, .01, and .001, respectively. Results are from Ordinary Least Squares (OLS) regression with robust standard errors (SE)}
\label{tab:irp_full_regressions_reciprocity}
\end{table*}

\begin{table*}[]
\centering
\resizebox{0.7\linewidth}{!}{%
\begin{tabular}{llll}
\hline
\multicolumn{1}{c}{}  &   & \multicolumn{2}{c}{Dependent Variables - IRP (De)Escalation Ratio}\\ \hline
\multicolumn{1}{c}{Dataset}  & \begin{tabular}[c]{@{}l@{}}Independent \\ Variables\end{tabular} & \multicolumn{1}{c}{\textit{Descalation}}  & \multicolumn{1}{c}{\textit{Escalation}} \\ \hline
 & CONST& \textit{49.00*** (0.000)}   & \textit{8.59 (0.312)} \\
 & SELF\_EXT& \textit{4.13** (0.002)} & \textit{0.46 (0.584)} \\
 & SELF\_AGR& \textit{1.91 (0.131)}& \textit{0.92 (0.125)} \\
 & SELF\_CON& \textit{-0.77 (0.499)}  & \textit{0.12 (0.850)} \\
 & SELF\_NEU& \textit{0.51 (0.644)}& \textit{-0.27 (0.684)}\\
 & SELF\_OPE& \textit{1.13 (0.287)}& \textit{0.09 (0.891)} \\
 & PARTNER\_EXT& \textit{2.85* (0.023)}  & \textit{0.36 (0.625)} \\
 & PARTNER\_AGR& \textit{1.66 (0.199)}& \textit{0.88 (0.139)} \\
 & PARTNER\_CON& \textit{-1.86 (0.084)}  & \textit{-0.19 (0.766)}\\
 & PARTNER\_NEU& \textit{-1.72 (0.167)}  & \textit{-0.33 (0.627)}\\
 & PARTNER\_OPE& \textit{-1.55 (0.161)}  & \textit{0.01 (0.982)} \\
\multirow{-12}{*}{Gemini 2.0 Flash} & POSITION & \textit{-19.33*** (0.000)}  & \textit{3.30** (0.003)}   \\
\rowcolor[HTML]{DEDCDC} 
\cellcolor[HTML]{DEDCDC}  & CONST& 96.69*** (0.000) & 19.60*** (0.001)   \\
\rowcolor[HTML]{DEDCDC} 
\cellcolor[HTML]{DEDCDC}  & SELF\_EXT& -0.41 (0.494) & 0.88 (0.056)\\
\rowcolor[HTML]{DEDCDC} 
\cellcolor[HTML]{DEDCDC}  & SELF\_AGR& 0.00 (1.000)  & -2.45*** (0.000)   \\
\rowcolor[HTML]{DEDCDC} 
\cellcolor[HTML]{DEDCDC}  & SELF\_CON& -0.64 (0.239) & -0.33 (0.447)  \\
\rowcolor[HTML]{DEDCDC} 
\cellcolor[HTML]{DEDCDC}  & SELF\_NEU& -0.14 (0.734) & 1.02* (0.024)  \\
\rowcolor[HTML]{DEDCDC} 
\cellcolor[HTML]{DEDCDC}  & SELF\_OPE& 0.10 (0.691)  & -0.25 (0.480)  \\
\rowcolor[HTML]{DEDCDC} 
\cellcolor[HTML]{DEDCDC}  & PARTNER\_EXT& -0.01 (0.984) & -0.22 (0.606)  \\
\rowcolor[HTML]{DEDCDC} 
\cellcolor[HTML]{DEDCDC}  & PARTNER\_AGR& -0.30 (0.512) & 0.06 (0.881)\\
\rowcolor[HTML]{DEDCDC} 
\cellcolor[HTML]{DEDCDC}  & PARTNER\_CON& 0.41 (0.498)  & -0.36 (0.413)  \\
\rowcolor[HTML]{DEDCDC} 
\cellcolor[HTML]{DEDCDC}  & PARTNER\_NEU& 1.37 (0.095)  & 0.47 (0.242)\\
\rowcolor[HTML]{DEDCDC} 
\cellcolor[HTML]{DEDCDC}  & PARTNER\_OPE& -0.01 (0.975) & -0.32 (0.405)  \\
\rowcolor[HTML]{DEDCDC} 
\multirow{-12}{*}{\cellcolor[HTML]{DEDCDC}Claude Sonnet 3.7} & POSITION & -1.05 (0.374) & 6.85*** (0.000)\\
 & CONST& 80.61*** (0.000) & 30.55*** (0.000)   \\
 & SELF\_EXT& 0.45 (0.318)  & 0.40 (0.428)\\
 & SELF\_AGR& 1.83*** (0.000)  & -1.37** (0.004)\\
 & SELF\_CON& 0.44 (0.301)  & 0.50 (0.290)\\
 & SELF\_NEU& -0.35 (0.386) & 0.70 (0.159)\\
 & SELF\_OPE& 0.05 (0.905)  & -0.04 (0.928)  \\
 & PARTNER\_EXT& -0.69 (0.158) & 1.56** (0.002) \\
 & PARTNER\_AGR& 0.64 (0.100)  & 0.11 (0.808)\\
 & PARTNER\_CON& -0.07 (0.867) & 0.01 (0.983)\\
 & PARTNER\_NEU& 0.20 (0.645)  & 0.70 (0.150)\\
 & PARTNER\_OPE& -0.01 (0.970) & -0.21 (0.639)  \\
\multirow{-12}{*}{GPT-4o mini}   & POSITION & -7.57*** (0.000) & 11.21*** (0.000)   \\
\rowcolor[HTML]{DEDCDC} 
\cellcolor[HTML]{DEDCDC}  & CONST& \cellcolor[HTML]{D9D9D9}105.01*** (0.000) & \cellcolor[HTML]{D9D9D9}28.92 (0.084)   \\
\rowcolor[HTML]{DEDCDC} 
\cellcolor[HTML]{DEDCDC}  & SELF\_EXT& \cellcolor[HTML]{D9D9D9}2.84 (0.134)  & \cellcolor[HTML]{D9D9D9}0.68 (0.603)\\
\rowcolor[HTML]{DEDCDC} 
\cellcolor[HTML]{DEDCDC}  & SELF\_AGR& \cellcolor[HTML]{D9D9D9}-0.35 (0.873) & \cellcolor[HTML]{D9D9D9}-0.73 (0.624)   \\
\rowcolor[HTML]{DEDCDC} 
\cellcolor[HTML]{DEDCDC}  & SELF\_CON& \cellcolor[HTML]{D9D9D9}-2.09 (0.284) & \cellcolor[HTML]{D9D9D9}-1.68 (0.284)   \\
\rowcolor[HTML]{DEDCDC} 
\cellcolor[HTML]{DEDCDC}  & SELF\_NEU& \cellcolor[HTML]{D9D9D9}-2.61 (0.146) & \cellcolor[HTML]{D9D9D9}1.01 (0.436)\\
\rowcolor[HTML]{DEDCDC} 
\cellcolor[HTML]{DEDCDC}  & SELF\_OPE& \cellcolor[HTML]{D9D9D9}1.33 (0.511)  & \cellcolor[HTML]{D9D9D9}0.56 (0.727)\\
\rowcolor[HTML]{DEDCDC} 
\cellcolor[HTML]{DEDCDC}  & PARTNER\_EXT& \cellcolor[HTML]{D9D9D9}-0.50 (0.748) & \cellcolor[HTML]{D9D9D9}0.16 (0.904)\\
\rowcolor[HTML]{DEDCDC} 
\cellcolor[HTML]{DEDCDC}  & PARTNER\_AGR& \cellcolor[HTML]{D9D9D9}0.27 (0.882)  & \cellcolor[HTML]{D9D9D9}1.42 (0.334)\\
\rowcolor[HTML]{DEDCDC} 
\cellcolor[HTML]{DEDCDC}  & PARTNER\_CON& \cellcolor[HTML]{D9D9D9}-1.37 (0.440) & \cellcolor[HTML]{D9D9D9}-2.40 (0.126)   \\
\rowcolor[HTML]{DEDCDC} 
\cellcolor[HTML]{DEDCDC}  & PARTNER\_NEU& \cellcolor[HTML]{D9D9D9}-0.10 (0.954) & \cellcolor[HTML]{D9D9D9}-0.13 (0.919)   \\
\rowcolor[HTML]{DEDCDC} 
\cellcolor[HTML]{DEDCDC}  & PARTNER\_OPE& \cellcolor[HTML]{D9D9D9}-1.34 (0.485) & \cellcolor[HTML]{D9D9D9}0.08 (0.956)\\
\rowcolor[HTML]{DEDCDC} 
\multirow{-12}{*}{\cellcolor[HTML]{DEDCDC}KODIS}  & POSITION & \cellcolor[HTML]{D9D9D9}0.44 (0.801)  & \cellcolor[HTML]{D9D9D9}8.25*** (0.000) \\\hline \cline{2-4} 
\end{tabular}}
\caption{Summary of all regression results for personality predictors across LLM (L2L) and human (KODIS) conversations, focusing on dependent variables related to escalation and de-escalation. Coefficients (B) are reported with p-values in parentheses. *, , *** indicate p $<$ .05, .01, and .001, respectively.}
\label{tab:escalation}
\end{table*}


\end{document}